\documentclass[sigconf]{acmart}
\AtBeginDocument{%
  \providecommand\BibTeX{{%
    \normalfont B\kern-0.5em{\scshape i\kern-0.25em b}\kern-0.8em\TeX}}}

\setcopyright{acmcopyright}
\copyrightyear{2023}
\acmYear{2023}
\acmDOI{10.1145/3581783.3613821}

\setcopyright{acmlicensed}\acmConference[MM '23]{Proceedings of the 31st
ACM International Conference on Multimedia}{October 29-November 3,
2023}{Ottawa, ON, Canada}
\acmBooktitle{Proceedings of the 31st ACM International Conference on
Multimedia (MM '23), October 29-November 3, 2023, Ottawa, ON, Canada}
\acmPrice{15.00}
\acmISBN{979-8-4007-0108-5/23/10}

\acmSubmissionID{2405}

\usepackage{multirow}
\usepackage{amsthm}
\usepackage{pifont}
\usepackage{subfigure}
\usepackage{algorithmic}
\usepackage{algorithm}
\usepackage{balance}

\def\ie{\emph{i.e.,~}}
\newcommand{\cmark}{\ding{51}}
\newcommand{\xmark}{\ding{55}}

\newtheorem{definition}{Definition}[section]
\newtheorem{assumption}{Assumption}[section]
\newtheorem{theorem}{Theorem}[section]
\newtheorem{claim}{Claim}[section]
\newtheorem{lemma}{Lemma}[section]
\newtheorem{remark}{Remark}

\begin{document}

\title{Chaos to Order: A Label Propagation Perspective on Source-Free Domain Adaptation}
\author{Chunwei Wu}
\affiliation{%
  \institution{East China Normal University}
  \city{Shanghai}
  \country{China}
}
\email{52215902005@stu.ecnu.edu.cn}

\author{Guitao Cao}
\authornote{Corresponding author.}
\affiliation{%
  \institution{Shanghai Key Laboratory of Trustworthy Computing, East China Normal University}
  \city{Shanghai}
  \country{China}
  \postcode{43017-6221}
}
\email{gtcao@sei.ecnu.edu.cn}

\author{Yan Li}
\affiliation{%
  \institution{Shanghai Normal University}
  \city{Shanghai}
  \country{China}
}
\email{yanli@shnu.edu.cn}

\author{Xidong Xi}
\affiliation{%
  \institution{East China Normal University}
  \city{Shanghai}
  \country{China}
}
\email{52265902004@stu.ecnu.edu.cn}

\author{Wenming Cao}
\affiliation{%
  \institution{Shenzhen University}
  \city{Shenzhen}
  \country{China}
}
\email{wmcao@szu.edu.cn}

\author{Hong Wang}
\affiliation{%
  \institution{Shanghai Research Institute of Microwave Equipment}
  \city{Shanghai}
  \country{China}
}
\email{wanghong1@cetc.com.cn}

\renewcommand{\shortauthors}{Chunwei Wu et al.}

\begin{abstract}
   Source-free domain adaptation (SFDA), where only a pre-trained source model is used to adapt to the target distribution, is a more general approach to achieving domain adaptation in the real world. 
   However, it can be challenging to capture the inherent structure of the target features accurately due to the lack of supervised information on the target domain. 
   By analyzing the clustering performance of the target features, we show that they still contain core features related to discriminative attributes but lack the collation of semantic information. 
   Inspired by this insight, we present \emph{Chaos to Order (CtO)}, a novel approach for SFDA that strives to constrain semantic credibility and propagate label information among target subpopulations. 
   CtO divides the target data into inner and outlier samples based on the adaptive threshold of the learning state, customizing the learning strategy to fit the data properties best. 
   Specifically, inner samples are utilized for learning intra-class structure thanks to their relatively well-clustered properties. 
   The low-density outlier samples are regularized by input consistency to achieve high accuracy with respect to the ground truth labels. 
   In CtO, by employing different learning strategies to propagate the labels from the inner local to outlier instances, it clusters the global samples from chaos to order. 
   We further adaptively regulate the neighborhood affinity of the inner samples to constrain the local semantic credibility. 
   In theoretical and empirical analyses, we demonstrate that our algorithm not only propagates from inner to outlier but also prevents local clustering from forming spurious clusters. 
   Empirical evidence demonstrates that CtO outperforms the state of the arts on three public benchmarks: Office-31, Office-Home, and VisDA.
\end{abstract}




\maketitle

\section{Introduction}
    
    \begin{figure}[t]
        \centering
        \includegraphics[width=8cm]{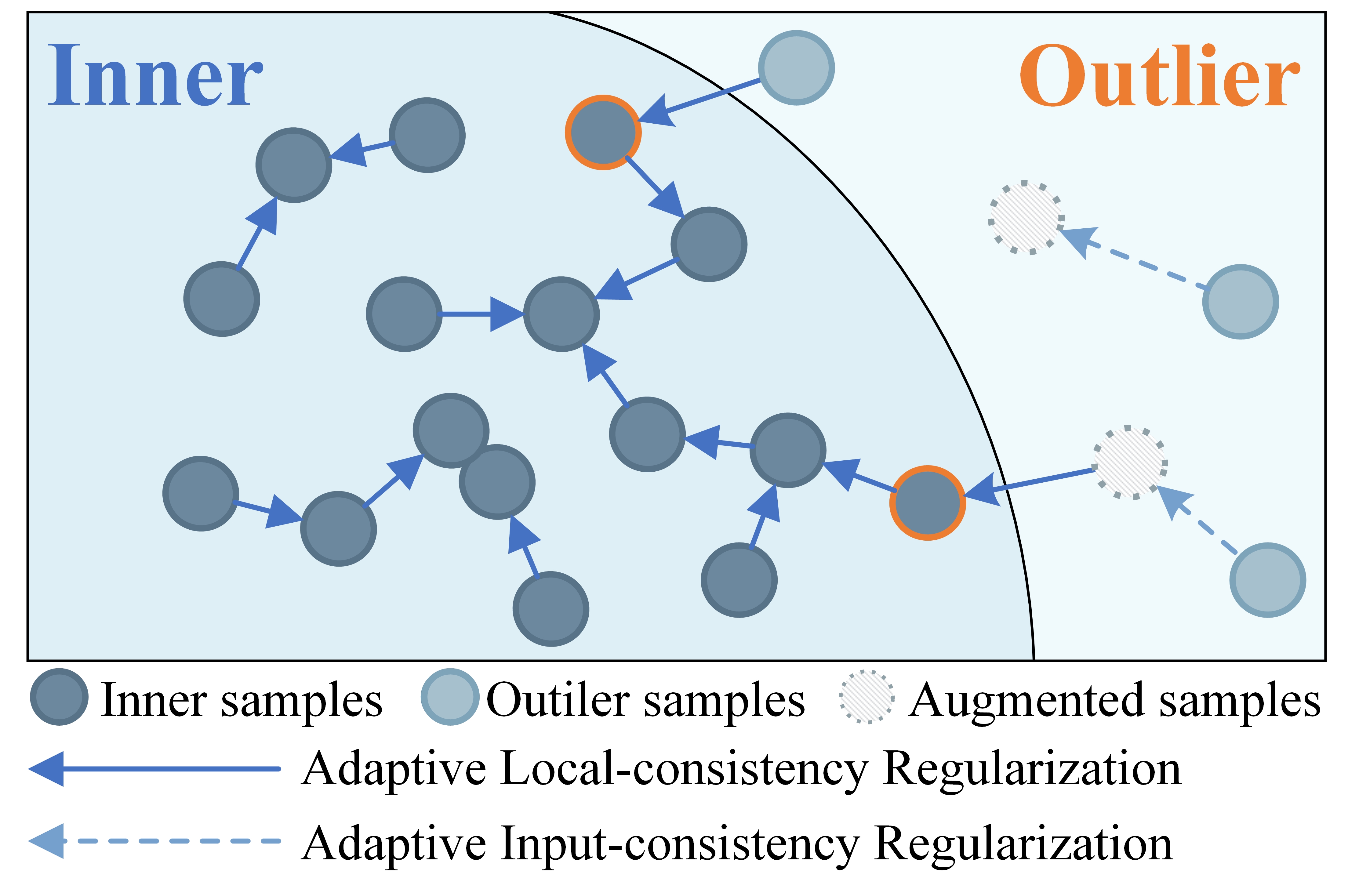}
        \caption{A toy illustration of our approach on label propagation. The target samples can be divided into two subsets: the inner set and the outlier set. CtO achieves label propagation through Adaptive Local-consistency Regularization and Adaptive Input-consistency Regularization. Among them, the outlier samples may move to the correct clusters with the help of their augmented versions. Maintaining this trend, the samples gradually move from chaos to order.}
        \label{vis:intro}
    \end{figure}
    
    The excellent performance of deep learning relies heavily on a large amount of high-quality labeled data. Obtaining large amounts of manually labeled data for specific learning tasks is often time-consuming and expensive, making these tasks challenging to implement in practical applications. To alleviate this dependency, Unsupervised Domain Adaptation (UDA) has been developed to improve performance in the unlabeled target domain by exploiting the labeled source domain. 
    Two popular practices for modern UDA design are learning domain-invariant features~\cite{DANN,CDAN,CAN,SRDC} and generating dummy samples to match the target domain distribution~\cite{DMRL,LiCWW20,E-MixNet,FixBi}. 
    However, due to data privacy and security issues, the source domain training data required by most existing UDA methods is usually unavailable in real-world applications. In response, Source-Free Domain Adaptation (SFDA) emerged, which attempted to adapt a trained source model to the target domain without using any source data.

    Due to the lack of source data, it is impossible to estimate source-target domain differences. Existing theoretical work usually provides learning guarantees on the target domain by further assuming that the source domain covers the support of the target domain. In the seminal work by~\cite{NRC}, the authors point out that the target features from the source model have formed some semantic structures. Inspired by this intuition, we can preserve the important clustering structure in the target domain by matching similar features in the high-dimensional space. However, the nearest-neighbor consistency of points in high-dimensional space may be wrong, such as when forcing the local consistency of points in low-density regions. 
    As shown in Table~\ref{tab:office-home_knn}, when the source and target domains have significant differences (\ie Pr$\rightarrow$Cl and Rw$\rightarrow$Cl), numerous features gather in low-density regions, with only about one-third of the neighbors having the correct labels.

    Along with such a question, we propose \emph{Chaos to Order (CtO)} (Fig.~\ref{vis:intro}), an effective method to achieve more robust clustering of unlabeled data from the perspective of label propagation.
    To achieve flexible adaptation for different data properties and exploit the target domain structure information, our work introduces a novel data division strategy and then designs different regularization strategies to achieve label propagation.

    Firstly, our approach treats the target domain's intrinsic structure information mining as a clustering problem. Although existing local consistency-based methods aim to preserve the local structure, Table~\ref{vis:intro} illustrates the reason why neighbors are unreliable: In distance-based neighbor discrimination, neighbors are similar points in a high-dimensional space, and since the points in the low-density region are all scattered far apart, the label information in the $K$-nearest neighbors is not consistent at this point. In CtO, we utilize the model's learning state to dynamically divide the target data into inner and outlier sets. The intrinsic reason is that a sample can be considered an inner sample if it can obtain high predictive values from the classifier; otherwise, it is an outlier. We regularize the input consistency of outliers and encourage local consistency for those inner samples, which effectively improves the mining of intrinsic structural information. 

    Secondly, we assume a minimum overlap in the subpopulations of the inner and outlier sets, and extend the subset using the simple but realistic extension assumption of~\cite{WeiSCM21}. For the inner set, the local-consistency regularizer connects similar points in the high-dimensional space, allowing SFDA training to proceed stably. 
    Enlightening experiments on Office-Home show that: (1) the pre-trained source model can extract rich semantic information from the target data; (2) what is lacking in domain adaptation is the filtering and permutation of high-dimensional semantic information. 
    Due to the lack of supervised information, we preserve the core features associated with the discriminative attributes by enforcing the local consistency of points in the latent space. 
    Moreover, we propose a re-weighted clustering strategy called adaptive Local-Consistency Regularization (ALR), which explicitly constrains the local semantic credibility to filter spurious clustering information.
    To advance further along this line, we propose Adaptive Input-Consistency Regularization (AIR) for the outlier set. 
    Generally, requiring the model to be invariant to input perturbations can improve generalizability. 
    Furthermore, as~\cite{WeiSCM21} discussed, a low-probability subset of data can be extended to a neighborhood with a large probability relative to that subset. 
    We show that labeling information can be propagated among subpopulations by minimizing the consistency regularization term on unlabeled data. In Theorem~\ref{thm:bound}, we give the upper bound on the task risk of the target model. 
    As a result, by customizing the learning strategy for different data properties, CtO can propagate structural information from the inner to the outlier while enhancing the clustering of the inner set.

    \begin{table}[t]
	\caption{Ratio (\%) of different number of $K$-nearest neighbor which have the correct predicted label (on Office-Home).}
        \label{tab:office-home_knn}
	\setlength\tabcolsep{3pt}
		\scalebox{0.9}{  
			\begin{tabular}{ccccccc}
				\toprule
				$K$ & Ar$\rightarrow$Cl & Ar$\rightarrow$Pr & Cl$\rightarrow$Ar & Pr$\rightarrow$Cl & Pr$\rightarrow$Rw & Rw$\rightarrow$Cl \\
				\midrule
				1 & 42.0 & 66.2 & 47.3 & 33.6 & 70.0 & 41.2 \\
				2 & 36.8 & 62.7 & 40.7 & 28.6 & 66.1 & 36.9 \\
				3 & 33.8 & 59.6 & 37.4 & 24.7 & 63.0 & 33.1 \\
				4 & 30.4 & 57.1 & 34.3 & 22.0 & 60.4 & 30.7 \\
				5 & 28.5 & 55.1 & 31.2 & 20.0 & 58.2 & 28.0 \\
				6 & 26.8 & 53.0 & 29.1 & 18.1 & 56.4 & 26.3 \\
				7 & 25.2 & 51.6 & 27.6 & 16.7 & 54.9 & 24.3 \\
				\bottomrule
			\end{tabular}
		}
    \end{table}

    In summary, the contributions of this paper are summarized as follows: 
    (1) We introduce CtO, a dynamical clustering approach for SFDA. Such a approach customizes the learning strategy for data subsets by using dynamic data splits, allowing label information to propagate among subpopulations. 
    (2) To combat spurious clustering, we propose a novel Adaptive Local-consistency Regularization (ALR) strategy that estimates ground-truth structural information by re-weighting the neighbors.
    (3) To utilize unlabeled data more effectively, we propose Adaptive Input-Consistent Regularization (AIR) from the perspective of label propagation. 
    By collaborating with ALR, structural information can be propagated from the inner to the outlier sets, significantly improving clustering performance. 
    (4) Empirical evidence demonstrates that the proposed method outperforms the state-of-the-art on three domain adaptation benchmark datasets.


    \begin{table*}[t]
        \caption{Comparison with different bottleneck layers on  \textbf{Office-Home}.}
        \label{tab:bottle_acc}
	\setlength\tabcolsep{3pt}
	\begin{center}
		\scalebox{0.85}{  
			\begin{tabular}{lccccccccccccc}
				\toprule
				Methods & Ar$\rightarrow$Cl & Ar$\rightarrow$Pr & Ar$\rightarrow$Rw & Cl$\rightarrow$Ar & Cl$\rightarrow$Pr & Cl$\rightarrow$Rw & Pr$\rightarrow$Ar & Pr$\rightarrow$Cl & Pr$\rightarrow$Rw & Rw$\rightarrow$Ar & Rw$\rightarrow$Cl & Rw$\rightarrow$Pr & Avg.\\
				\midrule
				AaD (w/ Source Bottleneck Layer) & 59.3 & 79.3 & 82.1 & 68.9 & 79.8 & 79.5 & 67.2 & 57.4 & 83.1 & 72.1 & 58.5 & 85.4 & 72.7 \\
				AaD (w/ Target Bottleneck Layer) & 69.3 & 85.7 & 91.4 & 82.4 & 86.2 & 87.4 & 84.5 & 67.5 & 90.5 & 89.1 & 68.9 & 92.1 & 82.9 \\
				\bottomrule
			\end{tabular}
		}
	\end{center}
    \end{table*}

\section{Related Work}

    \paragraph{Source-free Domain Adaptation (SFDA)}
        SFDA aims to adapt unlabeled target domains using only the pre-trained source model. Existing approaches try to refine the solution of SFDA by pseudo-labeling~\cite{SHOT,CPGA,HCL,SFDADE,BMD,JMDS}, generating transition domains~\cite{3CGAN,AAA,cPAE,featmixup}, or local consistency~\cite{GSFDA,NRC,AaD}. However, due to the domain differences, pseudo-labels that may contain noise cause confirmation bias. Additionally, task discriminative information and domain-related information are highly non-linearly entangled. Directly constructing an ideal generic domain from the source model may be difficult. Most closely related to our work is AaD\cite{AaD}, which introduced a simple and efficient optimization upper bound for feature clustering of unlabeled data, \ie aggregating (scattering) similar (dissimilar) features in the feature space. However, AaD uses $K$-nearest neighbors directly, which suffer from source bias due to domain shift. 
        In contrast to the above methods, we explore the idea of label propagation to assign regularization strategies to unlabeled data that are more suitable for the data properties, to achieve source-free model adaptation.

    \paragraph{Label Propagation}
        Label propagation has been widely used in semi-supervised learning. \cite{DouzeSHJ18} show that label propagation on large image sets outperforms state-of-the-art few-shot learning when few labels are available. \cite{IscenTAC19} employ a transductive label propagation method based on the stream shape assumption to predict the entire dataset. \cite{WeiSCM21} introduce the "extension" assumption to analyze label propagation and show learning guarantees for unsupervised and semi-supervised learning. \cite{CaiGLL21} extend the extension assumption to domain adaptation and propose a provably effective framework for domain adaptation based on label propagation.  
        Considering label propagation for SFDA and leveraging the advantages of extension assumptions, we design a novel and dynamic clustering strategy for SFDA that propagates structural information from high-density regions to low-density regions.

    \begin{figure}[t]
	\centering
	\includegraphics[width=7.5cm]{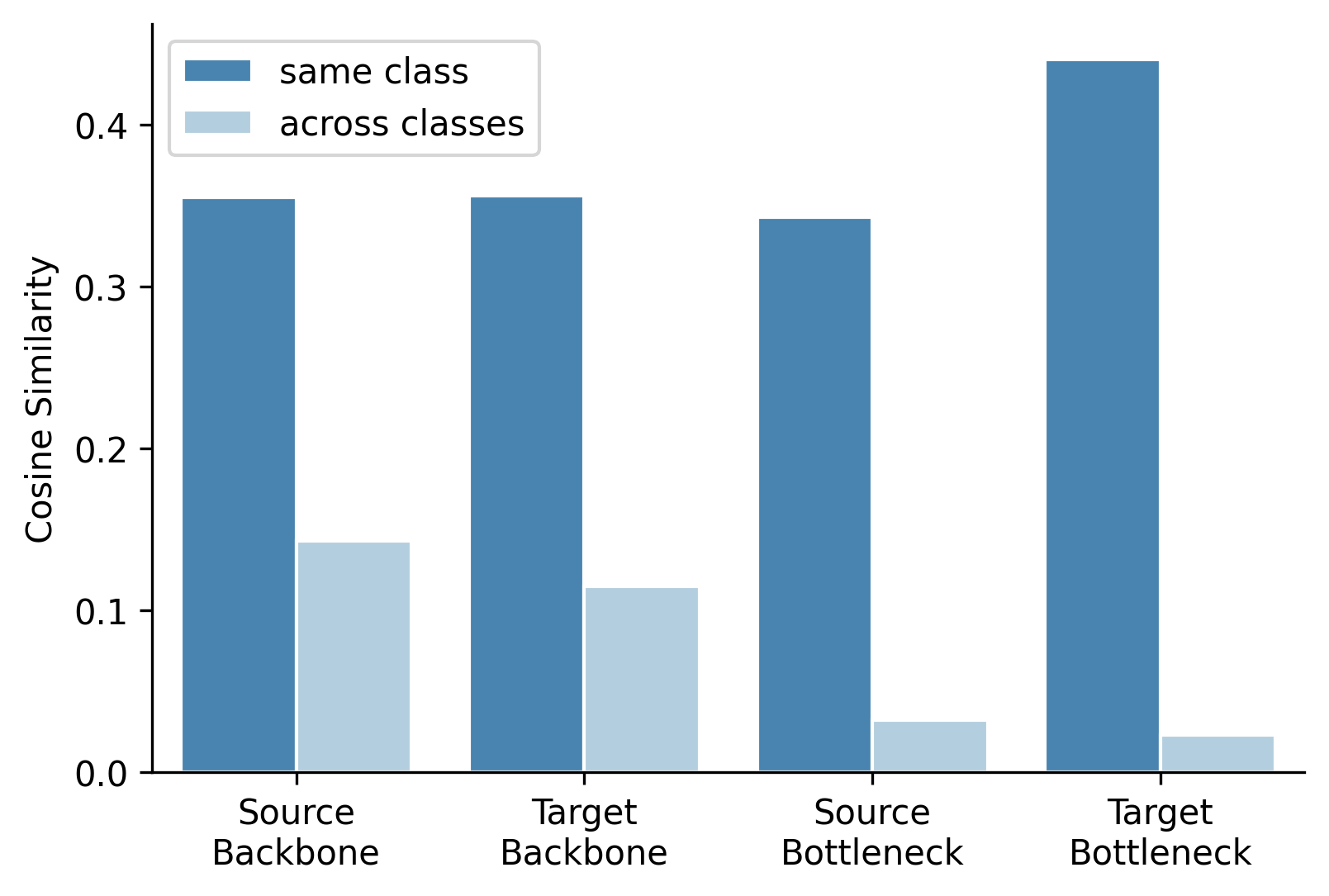}
	\caption{Cosine similarity within the same class and across classes on Office-Home.}
	\label{vis:sim}
    \end{figure}

\section{Preliminaries and Analysis}
\label{sec:PaA}

    In this section, we consider the clustering performance of the target domain and the label propagation for SFDA. 
    Correspondingly, we first introduce some notations and then perform an empirical and theoretical analysis to better understand the role of different learning strategies in CtO. 
    In Section~\ref{sec:ea}, we study an Oracle setup that beats the original AaD~\cite{AaD} by a large margin, confirming that the features from the source domain model are already rich in semantic information, which requires us to reduce the redundant information in the features. 
    Finally, in Section~\ref{sec:ta}, we claim that if the learning state of a model is superior, then the target sample has consistency with its neighbors (Claim~\ref{claim:dis_setting}). Furthermore, we present upper bounds on the target error in Theorem~\ref{thm:bound}.

    \subsection{Preliminary}
    For source-free domain adaptation (SFDA), consider an unlabeled target dataset $\mathcal{D}_T=\left\{x_i: x_i \in \mathcal{X}\right\}_{i=1}^{n_t}$ on the input space $\mathcal{X}_t$. The task is to adapt a well-trained source model to the target domain without source data, where the target domain has the same $|C|$ class as the source domain. 
    Following~\cite{NRC,AaD}, we use a feature extractor $h: \mathcal{X}_t \rightarrow \mathcal{Z}$, and the classifier $g_c: \mathcal{Z} \rightarrow \mathcal{C}$. Then the output of the network is denoted as $p(x)=\delta(g_c(h(x))) \in \mathcal{R}^C$, where $\delta$ is the softmax function.
    Specifically, we retrieve the nearest neighbors for each mini-batch of target features. Let $\boldsymbol{F} \in R^{n_t \times d}$ denotes a memory bank that stores all target features and $\boldsymbol{P} \in R^{n_t \times C}$ denotes the corresponding prediction scores in the memory bank, where $d$ is the feature dimension in the last linear layer: 
    \begin{equation}
	\begin{gathered}
		\boldsymbol{F}=\left[\boldsymbol{z}_1, \boldsymbol{z}_2, \ldots \boldsymbol{z}_{n_t}\right], \boldsymbol{P}=\left[\boldsymbol{p}_1, \boldsymbol{p}_2, \ldots \boldsymbol{p}_{n_t}\right],
	\end{gathered}
    \end{equation}
    where $\boldsymbol{z}_i$ is L2-normalized and $\boldsymbol{p}_i$ denotes the output softmax probability for $\boldsymbol{z}_i$.

    \subsection{Empirical analysis}
    \label{sec:ea}
    Most of the clustering-based SFDA methods have the problem of spurious clustering. Especially, in extreme domain shifts, the spurious clustering problem worsens. 
    To address this issue, we investigate the local consistency of feature representations on the source and target domain models. We carry out the experiments on Office-Home since it exists different degrees of domain shift, \ie Rw vs. Pr and Pr vs. Cl. 
    In this experiment, we study the feature properties of the different layers in the model: (1) Backbone: the last layer of the backbone network with 2048 feature dimensions; (2) Bottleneck: only replaces the bottleneck layer in the source model, with 256 feature dimensions. 
    
    It is worth noting that most of the existing clustering-based methods are distance-based. The key idea is the smoothness assumption that the model should produce similar predictions for similar unlabeled data. Therefore, a good feature representation should have intra-class compactness and inter-class separability. 
    Without loss of generality, we use cosine similarity as a metric. 
    It is very unexpected that the same-class similarity and across-class similarity between the source domain model and the target domain model on Backbone are similar, while a huge difference appears at the Bottleneck (see Fig.~\ref{vis:sim}). 
    According to the light blue bar in Fig.~\ref{vis:sim}, which visualizes the across-class similarity of samples at different network structures, we can easily observe that the features from the bottleneck layer have better inter-class separability.
    This means that adding a bottleneck layer to the model helps reduce redundant features, which improves discriminability and generalizability. 

    We further investigated the impact of bottleneck layers on the clustering-based SFDA approach by another study. Table~\ref{tab:bottle_acc} shows the learning effect of the AaD~\cite{AaD} with only the bottleneck layer replaced. Note that the bottleneck layer of the target model is only used for the analysis of this experiment. 
    We observe that replacing the target domain bottleneck layer improves the AaD model dramatically, from 72.7\% to 82.9\%.
    This indicates that the high-dimensional features from backbone network of the source model already contain rich semantic information, whereas the generalization of the features is more reflected in the filtering and permutation of the semantic information.
    Additionally, on the results of AaD (w/ Source Bottleneck Layer), there was a very strong correlation between prediction accuracy and the ratio of same-class similarity to across-class similarity, as indicated by the Spearman rank correlation of 0.92. 
    This observation hints that we can use the correlation between similarity and test accuracy to improve the clustering effect.

    We also evaluated the feature similarity of different metric functions. The different metric functions all point out that Backbone features are rich in semantic information (see Appendix~\ref{app:different_sim}).

    \subsection{Theoretical analysis}
    \label{sec:ta}

    Following the expansion assumption in~\cite{WeiSCM21,CaiGLL21}, we first define that the suitable set of input transformations $\mathcal{B}(\cdot)$ takes the general form $\mathcal{B}(x) \triangleq\{x^{\prime}: \exists A \in \mathcal{A}$ such that $\left\|x^{\prime}-A(x)\right\| \leq r \}$ for a small radius $r>0$, where $\mathcal{A}$ can be understood as a distance-based neighborhood or the data augmentations set. 
    Then, we define the neighborhood function $\mathcal{N}$ as
    \begin{equation}
	\mathcal{N}(x)= \left\{x^{\prime} \mid \mathcal{B}(x) \cap \mathcal{B}\left(x^{\prime}\right) \neq \emptyset\right\},
    \end{equation}
    and the neighborhood of a set $S \subset \mathcal{D}_T$ as 
    \begin{equation}
	\mathcal{N}(S) \triangleq \cup_{x \in S} \mathcal{N}(x).
    \end{equation}
    The regularizer of $G = g_c \circ h$ is defined as:
    \begin{equation}
	R_{\mathcal{B}}(G) = \mathbb{E}_{\mathcal{D}_T}\left[\max _{\text {neighbor } x^{\prime}} \mathbf{1}\left(G(x) \neq G(x^{\prime})\right)\right]
    \end{equation}

    Our setting for Source-free Domain Adaptation is formulated in the following assumption.
    \begin{assumption}\label{asm_setting}
	Let $I_i$ and $O_i$ denote the conditional distribution of the target distribution $\mathcal{D}_T$ on the set $I$ and $O$, respectively. Assume there exists a constant $\kappa \geq 1$ such that the measure $I_i$ and $O_i$ are bounded by $\kappa U_i$. That is, for any $S \subset \mathcal{X}_t$,
	$$P_{I_i}(S) \leq \kappa P_{U_i}(S)\text{ and }P_{O_i}(S) \leq \kappa P_{U_i}(S).$$
    \end{assumption}

    The expansion property on the target domain is defined as follows:
    \begin{definition}[Constant Expansion~\cite{WeiSCM21}] \label{def_expansion}
	We say that distribution $Q$ satisfies $(q, \xi)$-constant expansion for some constant $q, \xi \in (0, 1)$, if for all $S \subset Q$ satisfying $\mathbb{P}_{Q}(S) \ge q$, we have $\mathbb{P}_{Q}[\mathcal{N}(S) \backslash S] \ge \min \left\{\xi, \mathbb{P}_Q[S]\right\}$.
    \end{definition}

    Based on the model's learning state, our CtO method divides the target data into the inner set ($I$) and the outlier set ($O$). 
    The intuition lies in the fact that different datasets and classes should determine their division thresholds based on the models' learning states so that the division boundary is more reasonable. 
    Specifically, we set a global threshold $\rho$ to utilize unlabeled data at early training stages. As the adaptation progresses, we estimate the model’s learning state $\tau_i$ by the predictions of the model to determine a dynamic local (class-specific) threshold. 
    The following claim guarantees the consistency robustness of inner samples:
    \begin{claim}
    \label{claim:dis_setting}
	Suppose $G$ satisfies a Lipschitz condition; there exists a global threshold $\rho \in (0, 1)$ and scale of models' learning status $\tau_i$ such that the inner set $I$ is consistency robust, \ie $R_{\mathcal{B}}(G) = 0$. More specifically, 
        $$\begin{aligned}
            & r \leq (2 \max\{\tau_i\} \rho - 1) \frac{2}{L}, \\
            & \forall i \in [\mathcal{C}].
        \end{aligned}$$
    \end{claim}

    \begin{remark} The claim illustrates that when the global threshold $\rho$ is fixed, as long as the model's learning state is good enough, the inner set can achieve a consistency error close to zero. In clustering-based methods, $r$ usually refers to the number of selectable nearest neighbors, which allows us to select more reliable neighbor for inner samples. Moreover, we do not need to set specific data division thresholds for different datasets or classes based on experience, making it more general and realistic.
    \end{remark}

    With the above preparation, we can investigate how to propagate label information from the ordered inner set to the chaotic outlier set. 
    The following theorem establishes bounds on the target risks and indicates that as long as there is minimal overlap between the inner and outlier sets, label information will propagate in the target subpopulation. 
    
    \begin{theorem}
    \label{thm:bound}
	Suppose Assumption~\ref{asm_setting} and Claim~\ref{claim:dis_setting} hold and $I, O$ satisfies $(q, \mu)$-constant expansion. Then the expected error of model $G$ is bounded,
        $$\epsilon_{\mathcal{D}_T}(G) \leq 4 \max(q, \mu) \kappa + \mu (1 + \kappa).$$
    \end{theorem}
    
    \begin{remark} \label{rem:the}
    This theorem states that the target risks is bounded by the consistency regularization $R_{\mathcal{B}}$ (equivalently, $\mu$). Our analysis explains why consistency regularization is important for SFDA methods: assuming the data satisfies expansion, it encourages representations to maintain important semantic structures by enforcing local consistency within the representation space. 
    As a result, we can effectively capture the global structure of the target domain by considering all local neighborhoods together, which strongly enforces label propagation. 
    \end{remark}

    The proof of both claim and theorem are in Appendix~\ref{app:proof}.

\section{Method}
    In this section, we introduce CtO aims to achieve efficient feature clustering from the perspective of label propagation. 
    
    There are two aspects involved in achieving CtO. 
    First, \emph{how to divide the dataset reasonably?} The ideal inner set should have well-clustered properties, so the effectiveness of the division boundary in distinguishing high- and low-density regions is crucial when choosing a division threshold. The model's prediction probability can measure this. According to the previous analysis, by considering the model's learning state, our proposed adaptive division threshold can better guarantee local consistency among inner samples.
    
    Second, \emph{how can we customize learning strategies to achieve label propagation among target subpopulations?} Inspired by self-supervised learning, we apply two regularization strategies - local consistency and input consistency - on the inner and outlier sets, respectively. Our theoretical analysis demonstrates that these strategies can extend low-density subsets to high-density ones when the subpopulations overlap, which leads to the clustering from chaos to order.

    \subsection{Dynamic Data Grouping}
    As the analysis before, we employ the model's learning states to adaptively divide the data in $\mathcal{D}_T$ into the inner sets $I$ and outlier sets $O$. As believed in~\cite{FlexMatch}, the learning effect of the model can be reflected by the class-level hit rate. Therefore, our principle is that the data division in CtO should be related to the prediction confidence of the unlabeled data on different classes so as to reflect the class-level learning status. Namely, classes with fewer samples reaching a threshold of prediction confidence are considered to have difficult in learning local structural information. Moreover, the threshold should be increased steadily as the model is continuously improved during training. We set the global threshold as the exponential moving average (EMA) of the highest confidence level for each training time step:
    \begin{equation}
	\label{reweight:gc}
	\rho_t= \begin{cases}1/|C|, & \text { if } t=0 \\ \alpha \rho_{t-1}+(1-\alpha)  \max (p), & \text { otherwise }\end{cases}
    \end{equation}
    where $\alpha \in(0,1)$ is the momentum decay of EMA, $t$ denotes the $t$-th iteration. Combining this flexible thresholds, the learning effect of class $c$ in the time step is defined as: 
    \begin{equation}
	\tau_{t}(c)=\sum_{n=1}^{N_t} \mathbf{1}\left(\max \left(p \right)>\rho_t\right) \cdot \mathbf{1}\left(\arg \max \left(p=c\right)\right).
    \label{eq:le}
    \end{equation}

    Then we formulate the adaptive data division weights:
    \begin{equation}
	\begin{aligned}
	\label{reweight:div}
	&\mathcal{T}_t(c) =  \frac{1}{|C|} (1-\frac{\beta_t(c)}{\log \beta_t(c)}) \\
	&where,  \beta_t(c)=\frac{\tau_t(c)}{\max _c \tau_t}
	\end{aligned}
    \end{equation}

    Finally, the samples are dynamically grouped into the outlier set in the $t$-th iteration:
    \begin{equation}
		O^t=\left\{x_i \mid \max \left(p_i\right) \geq \mathcal{T}_t({\arg \max }\left(p_i\right)), x_i \in \mathcal{D}_T\right\} ,
    \label{eq:division}
    \end{equation}
    and the inner samples are the rest target data, \ie $I=\mathcal{D}_T \backslash O$.
    To this end, we customize learning strategies for different data properties and connect both sets by expansion assumption.
    
    \subsection{Label Propagation with Different Regularizations in SFDA}
    In the theoretical analysis (Section~\ref{sec:ta}), we show that performing consistency regularization on the unlabeled target data can propagate semantic information across different subpopulations. However, it also has some limitations. 
    First, the neighbors are not always correct. Due to the domain shift, the model may incorrectly focus on the object of some target features, leading to noisy labels. 
    Here, the misalignment of neighbors forms spurious clusters instead of helping label propagation. Second, when dealing with low-density regions (\ie outlier samples), designing the same local consistent regularization will further exacerbate the cross-label risk. 
    To address these limitations, we flexibly customize learning strategies for different data properties with the help of dynamic data grouping.
    
    \paragraph{Adaptive Local-consistency Regularization}
    In Adaptive Local-consistency Regularization (ALR), inspired by the fact that the target features from the source model have formed some semantic structures, we can capture the intra-class structure by local-consistency regularization. However, in the source-free domain adaptation problem, the features extracted by the pre-trained source model are typically influenced by the source bias. This may lead to neighbors containing incorrect semantic information. To mitigate incorrect alignment, we propose identifying the clustering weights of each sample.
    
    As observed in Fig.~\ref{vis:sim}, the cosine similarity of same-class is generally higher than that of across-class. Building on this finding, we can measure neighbor affinity based on cosine similarity and then re-weight the neighbors to approximate the ground-truth structural information. 
    By re-weighting with similarity-based adaptive weights, positive clustering can be promoted while spurious clustering can be combated. The Adaptive Local-consistency Regularization is as follows:
    \begin{equation}
	\label{loss:alr}
        \mathcal{L}_{alr} = -\sum_{i}^{N_{I}} \sum_{j}^{K_i} \boldsymbol{w}_{ij} \boldsymbol{p}_i^T \boldsymbol{p}_j
    \end{equation}
    where $K_i$ denotes the $K$-nearest neighbor set of $\boldsymbol{z}_i$. The similarity weight $\boldsymbol{w}_{ij}$ in Eq.~\ref{loss:alr} is the cosine similarity of $\boldsymbol{z}_i$ to the neighbors $\boldsymbol{z}_j$, which is calculated via the memory bank $\boldsymbol{F}$. 
    Optimizing $\mathcal{L}_{alr}$ improves the reliability of clustering, which stabilizes the intra-class structure. In addition, relaxing the ranking of samples in low-density regions helps reduce incorrect local semantic alignment.

    Additionally, to improve separability between clusters, we employ the separation strategy proposed by~\cite{AaD} to disperse the prediction of potentially dissimilar features. 
    \begin{equation}
	\label{loss:sep}
	\mathcal{L}_{sep} = -\sum_{i}^{N_{I}} \sum_{m}^{N_{B_i}} \boldsymbol{p}_i^T \boldsymbol{p}_m
    \end{equation}
    where $B_i$ denotes other features except $\boldsymbol{z}_i$ in mini-batch.

    \begin{table*}[t]
	\caption{Accuracy (\%) on \textbf{Office-Home} (ResNet-50).}
        \label{tab:office-home_sup}
	\setlength\tabcolsep{3pt}
	\begin{center}
		\scalebox{0.9}{  
			\begin{tabular}{lcccccccccccccc}
				\toprule
				Methods & Source-free & Ar$\rightarrow$Cl & Ar$\rightarrow$Pr & Ar$\rightarrow$Rw & Cl$\rightarrow$Ar & Cl$\rightarrow$Pr & Cl$\rightarrow$Rw & Pr$\rightarrow$Ar & Pr$\rightarrow$Cl & Pr$\rightarrow$Rw & Rw$\rightarrow$Ar & Rw$\rightarrow$Cl & Rw$\rightarrow$Pr & Avg.\\
				\midrule
				ResNet-50~\cite{ResNet} & \xmark & 34.9 & 50.0 & 58.0 & 37.4 & 41.9 & 46.2 & 38.5 & 31.2 & 60.4 & 53.9 & 41.2 & 59.9 & 46.1 \\
				CDAN~\cite{CDAN} & \xmark & 50.7 & 70.6 & 76.0 & 57.6 & 70.0 & 70.0 & 57.4 & 50.9 & 77.3 & 70.9 & 56.7 & 81.6 & 65.8 \\
				MDD~\cite{MDD} & \xmark & 54.9 & 73.7 & 77.8 & 60.0 & 71.4 & 71.8 & 61.2 & 53.6 & 78.1 & 72.5 & 60.2 & 82.3 & 68.1 \\
				SRDC~\cite{SRDC} & \xmark & 52.3 & 76.3 & 81.0 & 69.5 & 76.2 & 78.0 & 68.7 & 53.8 & 81.7 & 76.3 & 57.1 & 85.0 & 71.3 \\
				FixBi~\cite{FixBi} & \xmark & 58.1 & 77.3 & 80.4 & 67.7 & 79.5 & 78.1 & 65.8 & 57.9 & 81.7 & \textbf{76.4} & 62.9 & \textbf{86.7} & 72.7 \\
				\midrule
				SHOT~\cite{SHOT} & \cmark & 56.9 & 78.1 & 81.0 & 67.9 & 78.4 & 78.1 & 67.0 & 54.6 & 81.8 & 73.4 & 58.1 & 84.5 & 71.6 \\
				$A^2$Net~\cite{A2Net}  & \cmark & 58.4 & 79.0 & 82.4 & 67.5 & 79.3 & 78.9 & 68.0 & 56.2 & 82.9 & 74.1 & 60.5 & 85.0&  72.8 \\
				NRC~\cite{NRC}  & \cmark & 57.7& 80.3& 82.0& 68.1& 79.8& 78.6& 65.3& 56.4& 83.0& 71.0& 58.6& 85.6 & 72.2 \\
				SFDA-DE~\cite{SFDADE}  & \cmark & 59.7 & 79.5 & 82.4 & 69.7 & 78.6 & 79.2 & 66.1 & 57.2 & 82.6 & 73.9 & 60.8 & 85.5 & 72.9 \\
				feat-mixup~\cite{featmixup} & \cmark & \textbf{61.8} & 81.2 & 83.0 & 68.5 & 80.6 & 79.4 & 67.8 & \textbf{61.5} & \textbf{85.1} & 73.7 & \textbf{64.1} & 86.5 & 74.5 \\
				AaD~\cite{AaD} & \cmark & 59.3 & 79.3 & 82.1 & 68.9 & 79.8 & 79.5 & 67.2 & 57.4 & 83.1 & 72.1 & 58.5 & 85.4 & 72.7 \\
				DaC~\cite{DaC} & \cmark & 59.1 & 79.5 & 81.2 & 69.3 & 78.9 & 79.2 & 67.4 & 56.4 & 82.4 & 74.0 & 61.4 & 84.4 & 72.8 \\
                    NRC+ELR~\cite{NRC_ELR} & \cmark & 58.4 & 78.7 & 81.5 & 69.2 & 79.5 & 79.3 & 66.3 & 58.0 & 82.6 & 73.4 & 59.8 & 85.1 & 72.6 \\
                    SFUDA~\cite{UITR} & \cmark & 59.9 & \textbf{81.4} & 83.0 & 68.9 & 80.1 & 80.3 & 67.5 & 56.9 & 83.7 & 74.3 & 60.8 & 86.3 & 73.7 \\
				\midrule
				\textbf{CtO} & \cmark & 58.5 & 79.8 & \textbf{85.5} & \textbf{74.8} & \textbf{82.5} & \textbf{83.1} & \textbf{73.8} & 58.4 & 85.0 & \textbf{78.2} & 63.3 & \textbf{89.6} & \textbf{76.1} \\
				\bottomrule
			\end{tabular}
		}
	\end{center}
    \end{table*}

    \begin{table}[t]
	\caption{Accuracy (\%) on \textbf{Office-31} (ResNet-50).}
        \label{tab:office}
	\setlength\tabcolsep{1pt}
	\begin{center}
		\scalebox{0.85}{
			\begin{tabular}{lcccccccl}
				\toprule
				Methods & Source-free & A$\rightarrow$D & A$\rightarrow$W & D$\rightarrow$W & W$\rightarrow$D & D$\rightarrow$A & W$\rightarrow$A & Avg.\\
				\midrule
				ResNet-50~\cite{ResNet} & \xmark & 68.9 & 68.4 & 96.7 & 99.3 & 62.5 & 60.7 & 76.1 \\
				CDAN~\cite{CDAN} & \xmark & 92.9 & 94.1 & 98.6 & 100.0 & 71.0 & 69.3 & 87.7 \\
				MDD~\cite{MDD} & \xmark & 90.4 & 90.4 & 98.7 & 99.9 & 75.0 & 73.7 & 88.0 \\
				SRDC~\cite{SRDC} & \xmark & 95.8 & 95.7 & 99.2 & 100.0 & 76.7 & 77.1 & 90.8 \\
				FixBi~\cite{FixBi} & \xmark & 95.0 & 96.1 & 99.3 & 100.0 & 78.7 & 79.4 & 91.4 \\
				\midrule
				SHOT~\cite{SHOT} & \cmark & 93.1 & 90.9 & 98.8 & 99.9 & 74.5 & 74.8 & 88.7 \\
				$A^2$Net~\cite{A2Net} & \cmark & 94.5 & 94.0 & 99.2 & \textbf{100.0} & 76.7 & 76.1 & 90.1 \\
				NRC~\cite{NRC} & \cmark & 96.0 & 90.8 & 99.0 & \textbf{100.0} & 75.3 & 75.0 & 89.4 \\
				HCL~\cite{HCL} & \cmark & 94.7 & 92.5 & 98.2 & \textbf{100.0} & 75.9 & 77.7 & 89.8 \\
				SFDA-DE~\cite{SFDADE} & \cmark & 96.0 & 94.2 & 98.5 & 99.8 & 76.6 & 75.5 & 90.1 \\
				AaD~\cite{AaD} & \cmark & \textbf{96.4} & 92.1 & \textbf{99.1} & \textbf{100.0} & 75.0 & 76.5 & 89.9 \\
				feat-mixup~\cite{featmixup} & \cmark & 94.6 & 93.2 & 98.9 & \textbf{100.0} & 78.3 & 78.9 & 90.7 \\
                    NRC+ELR~\cite{NRC_ELR} & \cmark & 93.8 & 93.3 & 98.0 & \textbf{100.0} & 76.2 & 76.9 & 89.6 \\
                    SFUDA~\cite{UITR} & \cmark & 96.2 & 94.0 & \textbf{99.1} & 99.9 & 76.9 & \textbf{79.2} & 90.7 \\
				\midrule
				\textbf{CtO} & \cmark  & \textbf{96.4} & \textbf{95.1} & 99.0 & \textbf{100.0} & \textbf{80.0} & 78.2 & \textbf{91.5} \\
				\bottomrule
			\end{tabular}
		}
	\end{center}
    \end{table}
    
    \paragraph{Adaptive Input-consistency Regularization}
    In Adaptive Input-consistency Regularization, we propagate the structural information from the inner set to the outlier set as discussed in Remark~\ref{rem:the}. Since the outliers in the low-density region are far away from all other points, which means there is no nearest neighbor support, we turn to seek support from the outliers themselves. Specifically, we use a weakly augmented version of $x_i$, denoted as $\omega(x_i)$, to generate the pseudo-label $\hat{p}_i = P(y\mid\omega(x_i))$ and enforce consistency against its strongly augmented version $\Omega(x_i)$. To encourage the model to make diverse predictions, we combined regularization with the aforementioned class-level confidence thresholds.
    The Adaptive Input-consistency Regularization is as follows:
    \begin{equation}
	\label{loss:air}
	\mathcal{L}_{air} = \frac{1}{N_{O}} \sum_{i=1}^{N_{O}} \mathcal{H}(\hat{p}_i, q_i),
    \end{equation}
    where $\mathcal{H}(\cdot, \cdot)$ refers to cross-entropy loss, and $q_i = P(y\mid\Omega(x_i))$ denotes the pseudo label of $\Omega(x_i)$.

    During training, the clustering processes on the inner and outlier sets facilitate each other. By implementing different regularization strategies, labels are propagated among subsets to enable under- or hard-to-learn samples to find suitable neighbors. The inclusion of these new members in clusters provides additional information for learning the intra-class structure, which adjusts the feature space and enhances the power of clustering.

    \subsection{Overall Objective}
    As described above, the overall objective of CtO can be summarized as follows:
    \begin{equation}
	\label{loss:totle_loss}
	\mathcal{L} = \mathcal{L}_{alr} + \mathcal{L}_{air} +  \lambda \mathcal{L}_{sep},
    \end{equation}
    where $\lambda$ are a trade-off parameter. 
    With $\mathcal{L}_{alr}$ and $\mathcal{L}_{air}$, CtO preserves local and input consistency, allowing label information to be propagated. The training process is described in Appendix~\ref{app:CtO}.

    \begin{table*}[t]
	\caption{Accuracy (\%) on \textbf{VisDA} (ResNet-101).}
        \label{tab:visda}
	\setlength\tabcolsep{3pt}
	\begin{center}
		\scalebox{1.}{
			\begin{tabular}{lcccccccccccccc}
				\toprule
				Methods & Source-free & plane & bicycle & bus & car & horse & knife & mcycl & person & plant & sktbrd & train & truck & Per-class\\
				\midrule
				ResNet-101~\cite{ResNet} & \xmark & 55.1 & 53.3 & 61.9 & 59.1 & 80.6 & 17.9 & 79.7 & 31.2 & 81.0 & 26.5 & 73.5 & 8.5 & 52.4 \\
				CDAN~\cite{CDAN} & \xmark & 85.2 & 66.9 & 83.0 & 50.8 & 84.2 & 74.9 & 88.1 & 74.5 & 83.4 & 76.0 & 81.9 & 38.0 & 73.9 \\
				SAFN~\cite{SAFN} & \xmark & 93.6 & 61.3 & 84.1 & 70.6 & 94.1 & 79.0 & 91.8 & 79.6 & 89.9 & 55.6 & 89.0 & 24.4 & 76.1 \\
				MCC~\cite{MCC} & \xmark & 88.7 & 80.3 & 80.5 & 71.5 & 90.1 & 93.2 & 85.0 & 71.6 & 89.4 & 73.8 & 85.0 & 36.9 & 78.8 \\
				FixBi~\cite{FixBi} & \xmark & 96.1 & 87.8 & 90.5 & 90.3 & 96.8 & 95.3 & 92.8 & 88.7 & 97.2 & 94.2 & 90.9 & 25.7 & 87.2 \\
				\midrule
				SHOT~\cite{SHOT} & \cmark & 92.6 & 81.1 & 80.1 & 58.5 & 89.7 & 86.1 & 81.5 & 77.8 & 89.5 & 84.9 & 84.3 & 49.3 & 79.6 \\
				$A^2$Net~\cite{A2Net} & \cmark & 94.0 & 87.8 & 85.6 & 66.8 & 93.7 & 95.1 & 85.8 & 81.2 & 91.6 & 88.2 & 86.5 & 56.0 & 84.3 \\
				NRC~\cite{NRC} & \cmark & 96.8 & \textbf{91.3} & 82.4 & 62.4 & 96.2 & 95.9 & 86.1 & 80.6 & 94.8 & 94.1 & 90.4 & 59.7 & 85.9 \\
				HCL~\cite{HCL} & \cmark & 93.3 & 85.4 & 80.7 & 68.5 & 91.0 & 88.1 & 86.0 & 78.6 & 86.6 & 88.8 & 80.0 & \textbf{74.7} & 83.5 \\
				CPGA~\cite{CPGA} & \cmark & 94.8 & 83.6 & 79.7 & 65.1 & 92.5 & 94.7 & 90.1 & 82.4 & 88.8 & 88.0 & 88.9 & 60.1 & 84.1 \\
				SFDA-DE~\cite{SFDADE} & \cmark & 95.3 & 91.2 & 77.5 & 72.1 & 95.7 & 97.8 & 85.5 & \textbf{86.1} & 95.5 & 93.0 & 86.3 & 61.6 & 86.5 \\
				AaD~\cite{AaD} & \cmark & 97.4 & 90.5 & 80.8 & 76.2 & 97.3 & 96.1 & 89.8 & 82.9 & 95.5 & 93.0 & 92.0 & 64.7 & 88.0 \\
				DaC~\cite{DaC} & \cmark & 96.6 & 86.8 & \textbf{86.4} & \textbf{78.4} & 96.4 & 96.2 & \textbf{93.6} & 83.8 & \textbf{96.8} & \textbf{95.1} & 89.6 & 50.0 & 87.3 \\
				\midrule
				\textbf{CtO} & \cmark & \textbf{98.2} & 91.0 & \textbf{86.4} & 78.0 & \textbf{97.6} & \textbf{98.8} & 91.8 & 84.8 & 96.6 & 94.7 & \textbf{93.7} & 53.3 & \textbf{88.7} \\
				\bottomrule
			\end{tabular}
		}
	\end{center}
    \end{table*}

    \begin{table}[t]
        \caption{Ablation study on Office-31.}
        \label{tab:ablation}
        \begin{center}
    \scalebox{0.8}{
        \begin{tabular}{cccc|ccccc}
            \toprule
            $\mathcal{L}_{seq}$ & $\mathcal{L}_{lr}$ & $\mathcal{L}_{alr}$& $\mathcal{L}_{air}$ & A$\rightarrow$D & A$\rightarrow$W & D$\rightarrow$A & W$\rightarrow$A & Avg. \\
            \midrule
            \cmark & \cmark & & & \textbf{96.4} & 92.1 & 75.0 & 76.5 & 85.0 \\
            \cmark & & \cmark & & 95.4 & 93.3 & 77.9  & 77.6 & 86.1 \\
            \cmark & &  & \cmark & 95.8 & 94.7 & 79.4 & 77.8 & 86.9 \\
            \cmark & & \cmark & \cmark & \textbf{96.4} & \textbf{95.1} & \textbf{80.0} & 78.2 & \textbf{87.4} \\
            \bottomrule
        \end{tabular}
    }
        \end{center}
    \end{table}

\section{Experiments}
\label{sec:exp}

    \subsection{Setup}
        \paragraph{Datasets}
        We conduct experiments on three public domain adaptation benchmarks. 
        (i) \textbf{Office-31}~\cite{Office_31} is a commonly used dataset for domain adaptation that consists of three domains: Amazon (A), Webcam (W), and DSLR (D), each containing 31 categories of items in an office environment.
        (ii) \textbf{Office-Home}~\cite{Office_Home} is a standard domain adaptation dataset collected in office and home environments. 
        It consists of four domains, Art (Ar), Clipart (Cl), Product (Pr), and RealWorld (Rw), and each covering 65 object categories. 
        (iii) \textbf{VisDA}~\cite{VisDA} is one of the large benchmark datasets on the domain adaptation task. It contains 12 categories of images from two subsets: synthetic image domain and real image domain.

        \paragraph{Implementation details}
        Following the standard protocol for SFDA, we use all labeled source data to obtain pre-trained models. 
        For the Office-31 and Office-Home, the backbone network is ResNet-50~\cite{ResNet}. For VisDA, the backbone network is ResNet-101. 
        For a fair comparison, we use the same network structure as SHOT~\cite{SHOT}, NRC~\cite{NRC} and AaD~\cite{AaD}. 
        All network parameters are updated by Stochastic Gradient Descent (SGD) with momentum of 0.9, an initial learning rate of 0.001, and a weight decay of 0.005. The learning rate of the additional layer is 10 times smaller than that of the backbone layer. 
        We follow G-SFDA~\cite{GSFDA}, NRC~\cite{NRC}, and AaD~\cite{AaD} for the number of nearest neighbors ($K$): set 3 for Office-31, Office-Home, and 5 on VisDA.
        To ensure a fair comparison, we set the hyperparameter $\lambda$ to be the same as in the previous work~\cite{AaD}. That is, we set $\lambda=\left(1+10 * \frac{iter}{max\_iter}\right)^{-\beta}$, and set $\beta$ to 0 on Office-Home, 2 on Office-31, and 5 on VisDA. 
        The strong augmentation function used in our experiments is RandAugment \cite{RandAugment}.

        \begin{figure*}[t]
            \centering
            \begin{minipage}{0.23\linewidth}
                \subfigure[Source-only (A$\rightarrow$W)]{
                    \includegraphics[width=4cm]{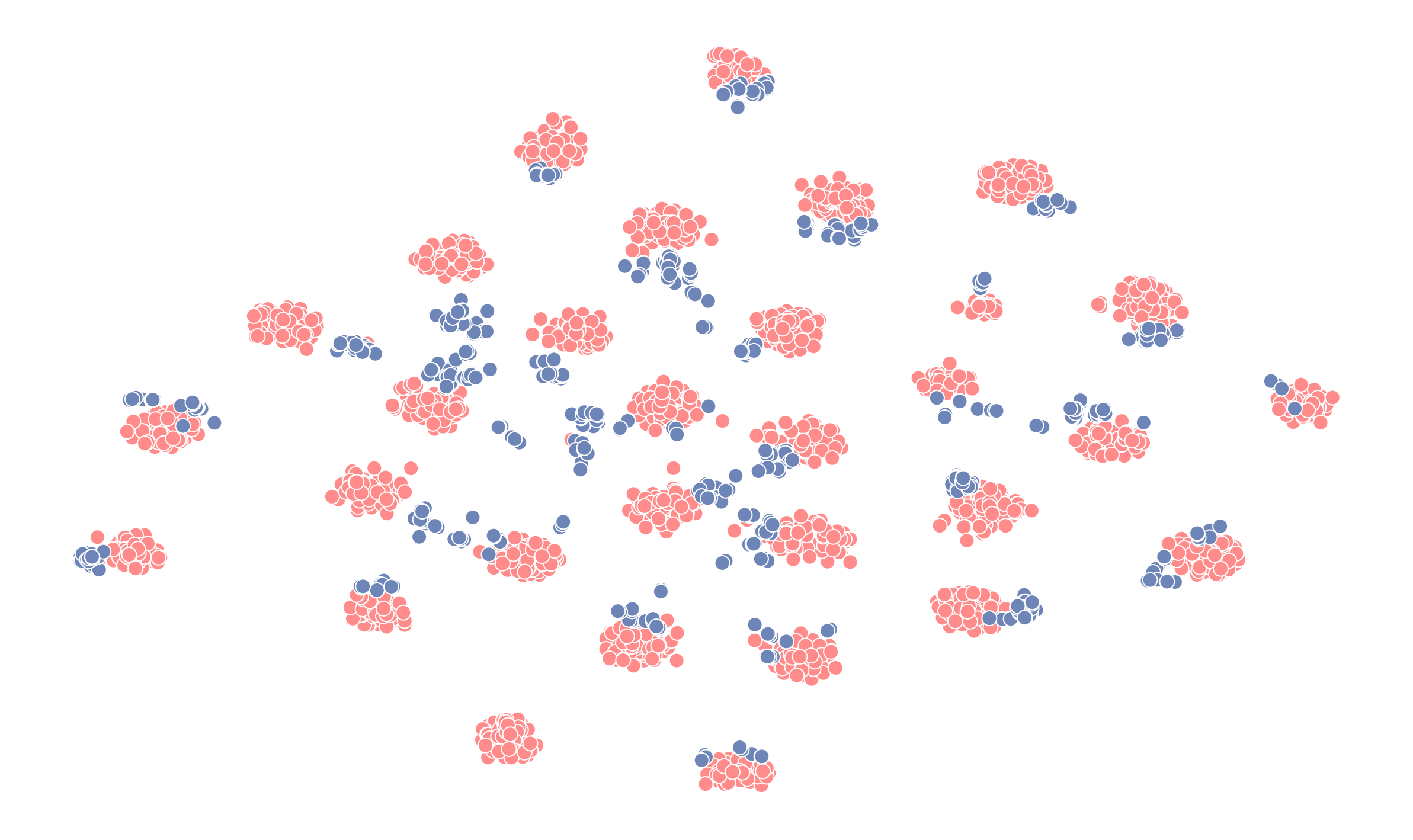}
                    \label{vis:a_w_res}
                }
            \end{minipage}
            \begin{minipage}{0.23\linewidth}
                \subfigure[CtO (A$\rightarrow$W)]{
                    \includegraphics[width=4cm]{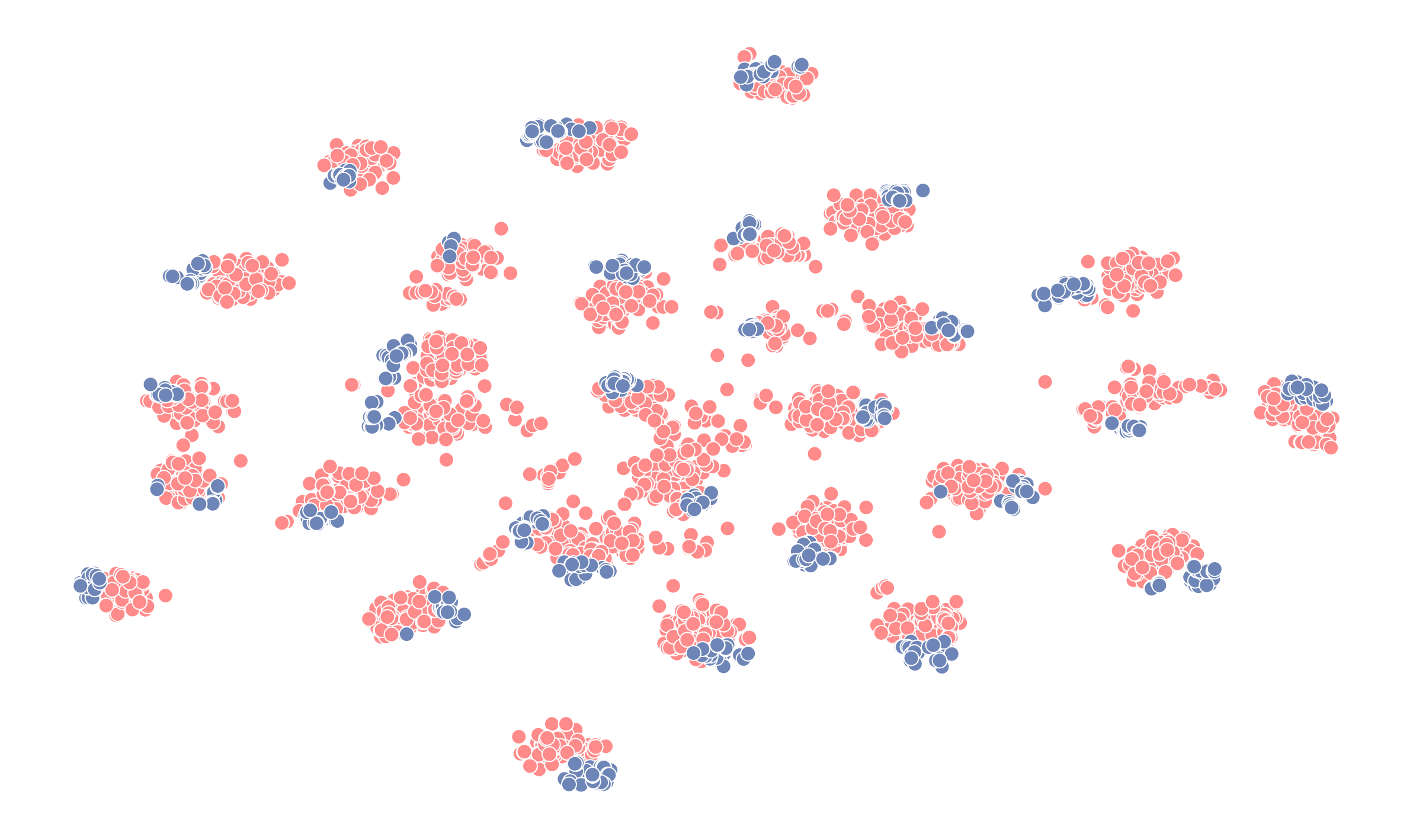}
                    \label{vis:a_w_alt}
                }
            \end{minipage}
            \begin{minipage}{0.23\linewidth}
                \subfigure[Source-only (D$\rightarrow$A)]{
                    \includegraphics[width=4cm]{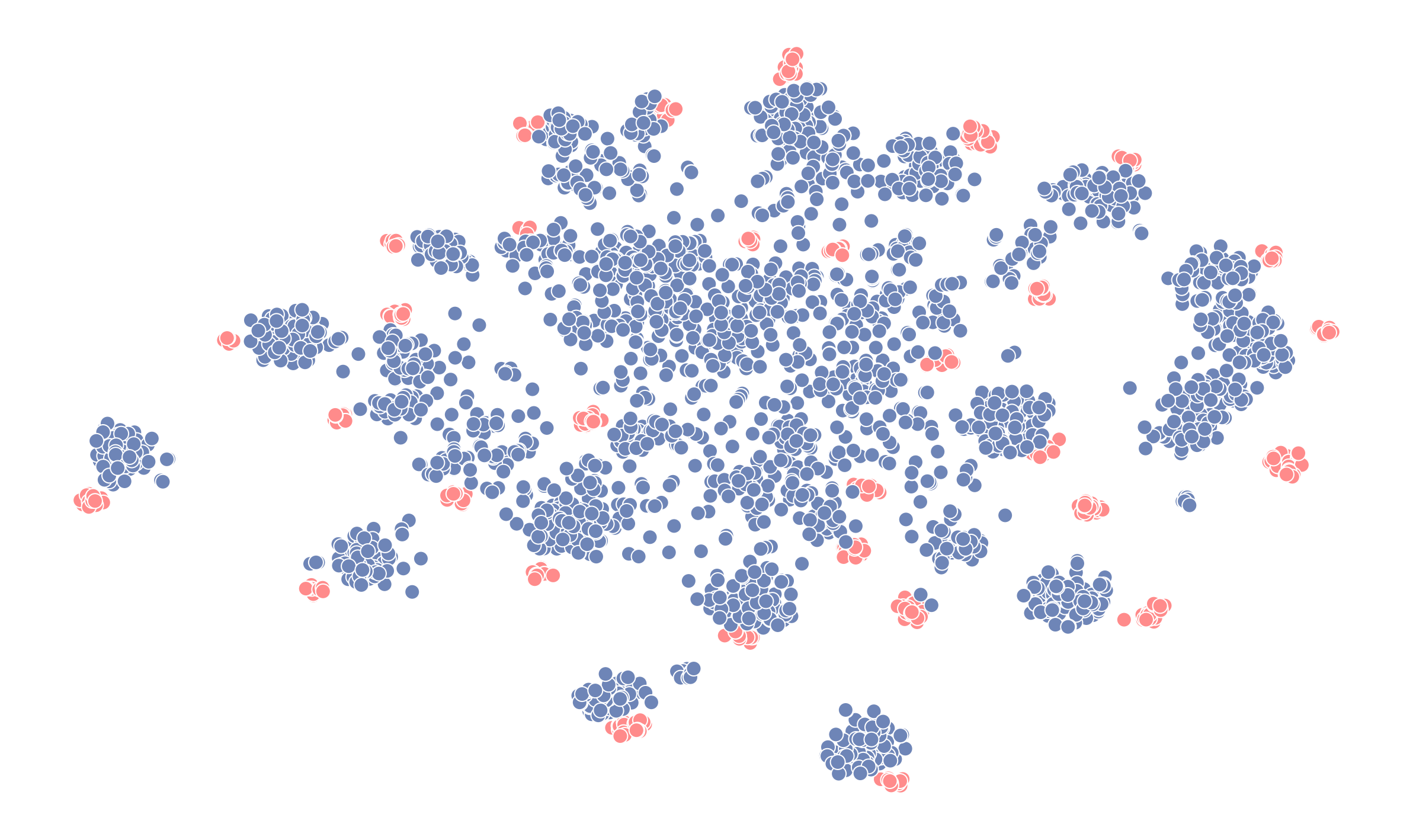}
                    \label{vis:d_a_res}
                }
            \end{minipage}
            \begin{minipage}{0.23\linewidth}
                \subfigure[CtO (D$\rightarrow$A)]{
                    \includegraphics[width=4cm]{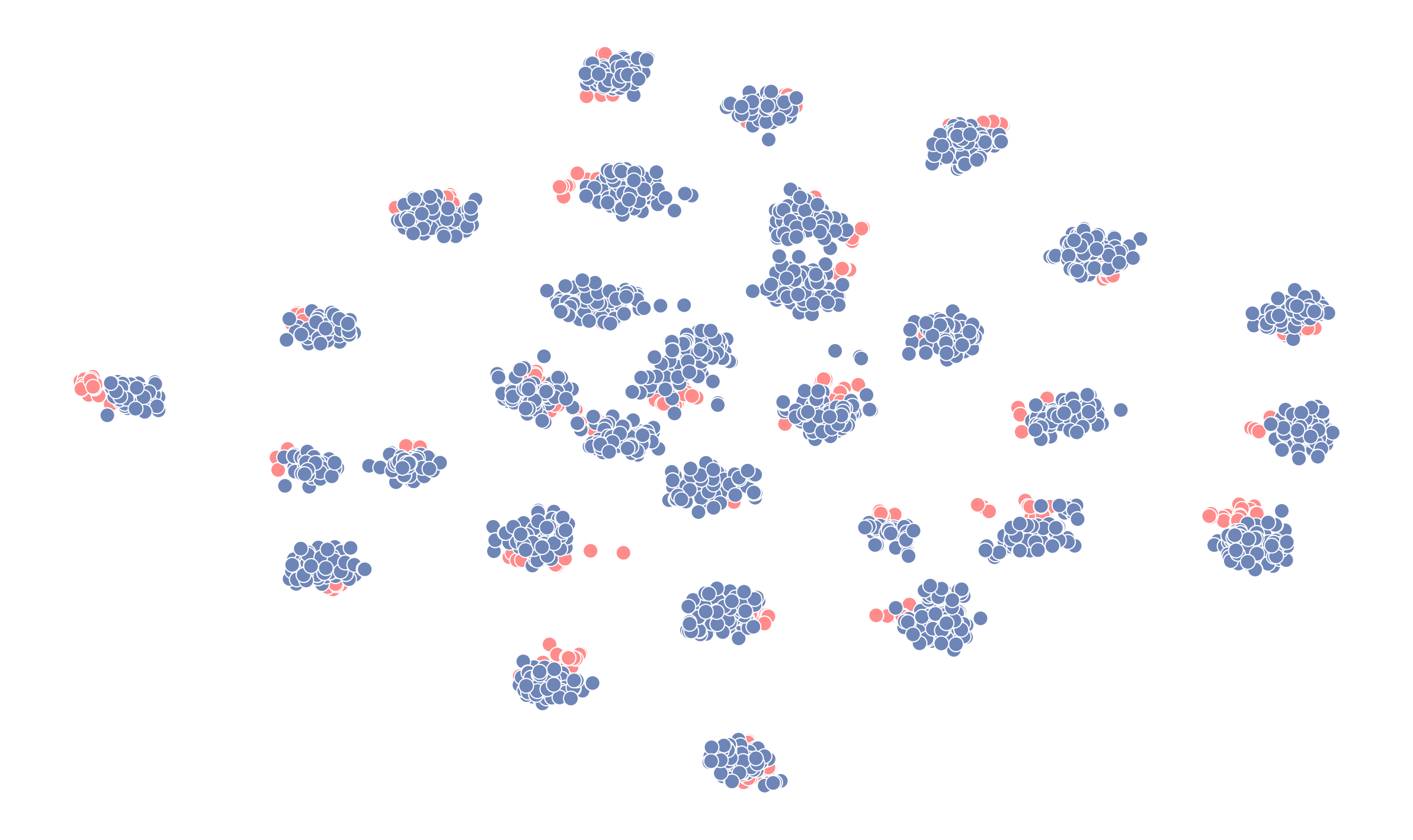}
                    \label{vis:d_a_alt}
                }
            \end{minipage}
            
            \begin{minipage}{0.46\linewidth}
                \subfigure[Source-only (D$\rightarrow$A)]{
                    \includegraphics[width=8cm]{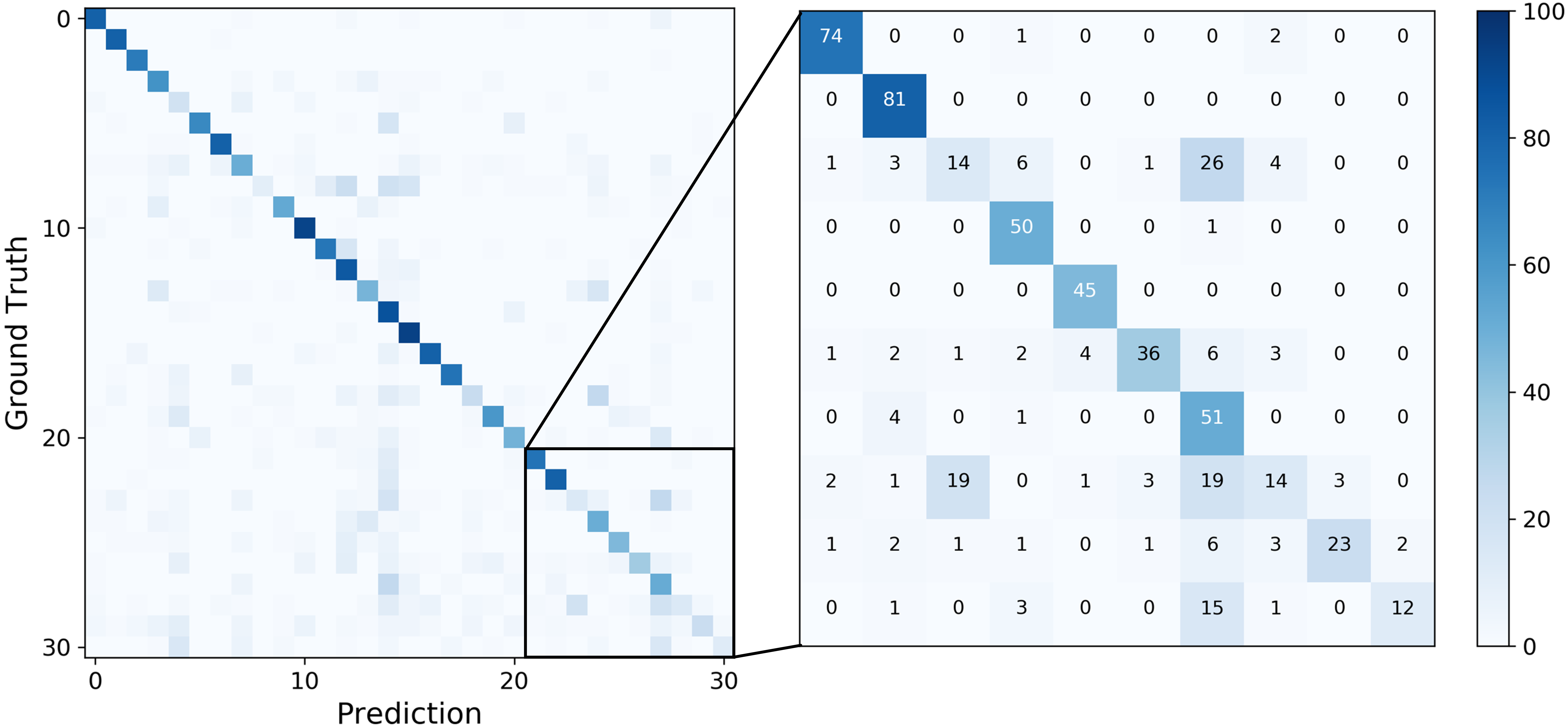}
                    \label{vis:d_a_con_res}
                }
            \end{minipage}
            \begin{minipage}{0.46\linewidth}
                \subfigure[CtO (D$\rightarrow$A)]{
                    \includegraphics[width=8cm]{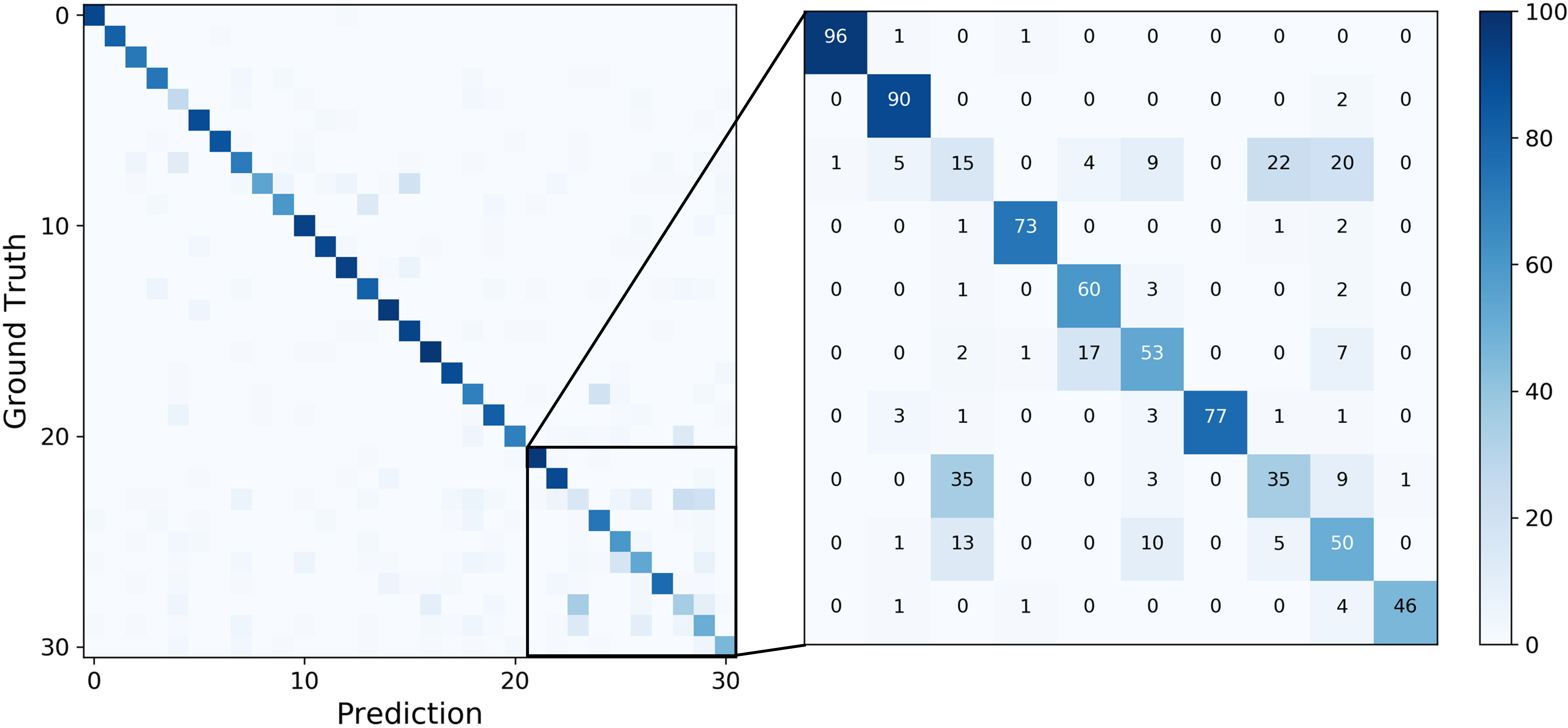}
                    \label{vis:d_a_con_alt}
                }
            \end{minipage}
            \caption{The t-SNE and Confusion Matrix visualization. (a-d): t-SNE visualization of the final prediction layer activation for source model and CtO, where red and blue points denote the source and target domains, respectively. Note that the source samples are only used to plot the t-SNE. (e) and (f): Confusion Matrix visualization for source model and CtO.}
            \label{vis:tsne_con}
        \end{figure*}

    \subsection{Results and Analysis}
        In this section, we will present our results and compare with other methods, which are summarized in Table~\ref{tab:office-home_sup},~\ref{tab:office},~\ref{tab:visda}, respectively. For a fair comparison, all baseline results were obtained from their original papers or the follow-up work.

        \paragraph{Comparison with state-of-the-art methods}
        For Office-31, as shown in Table~\ref{tab:office}, the proposed CtO yield state-of-the-art performance on 4 out of 6 tasks.
        Note that our CtO produces competitive results even when compared to source-present methods such as FixBi~\cite{FixBi} (91.5\% v.s. 91.4\%). 
        For Office-Home, Table~\ref{tab:office-home_sup} presents that the proposed CtO method achieves the most advanced classification accuracy (76.1\%) and achieves the highest results on 7 out of 12 tasks. 
        As we all know, in clustering-based methods, the clustering error increases with the number of object classes. Therefore, it is difficult for local consistency-based SFDA methods to accurately capture the target structure information. However, our CtO employs adaptive input-consistency regularization to efficiently utilize unlabeled data through label propagation. This is the primary reason for our success on Office-Home. Moreover, CtO beats several source-present DA methods, such as SRDC~\cite{SRDC} and FixBi~\cite{FixBi}, by a large margin, which means that even if we do not have access to the source data, our method can still exploit the target structure information to achieve better adaptation. 
        Similar observations on VisDA can be found in Table~\ref{tab:visda}. The reported results sufficiently demonstrate the superiority of our method.

        \paragraph{Comparison with clustering-based Method}
        As discussed in related work, NRC~\cite{NRC} uses reciprocal nearest neighbors to measure clustering affinity. 
        On the hard tasks of Office-Home, our approach outperforms the accuracy of NRC by a considerable margin, especially on tasks Pr$\rightarrow$Ar (73.8\% v.s. 65.3\%). 
        The improvement of our method indicates the importance of our adaptive input-consistency regularization for capturing intra-class structural information. 
        Compared with AaD~\cite{AaD}, our CtO improves the accuracy by 1.6\% on Office-31 and by 3.1\% on Office-Home, indicating that the co-training of the adaptive local-consistency regularizer and the adaptive input-consistency regularizer performs reliable label propagation through the subpopulation of unlabeled data. 
        Moreover, in VisDA, CtO exhibits higher recognition accuracy for several confusing objects than AaD, which indicates that the adaptive input-consistency regularizer can enhance model discrimination by providing more comprehensive intra-class information.

        \paragraph{Ablation Study}
        To evaluate the contribution of the different components of our work, we conduct ablation studies for CtO on Office-31. We investigated different combinations of the two parts: Adaptive Local-consistency Regularization (ALR) and Adaptive Input-consistency Regularization (AIR). Compared to our method, AaD (\ie only $\mathcal{L}_{seq}$ and $\mathcal{L}_{lr}$ are used) can be regarded as the baseline. As shown in Table~\ref{tab:ablation}, each part of our method contributes to improving performance. 
        It is not difficult to find that AIR contributes the most to the improvement of accuracy, with the performance increasing from 85.0\% to 86.9\%, which shows the effectiveness of label propagation. 
        ALR also improves the average performance by 1.1\% compared to the base model, confirming that the distance-based re-weighting improves the quality of the neighbors. 
        For easy transfer tasks, target features from pre-trained source models naturally have good clustering performance. In this case, ALR dominates in loss optimization, with AIR helping to improve model training for under-learned categories. 
        When the target feature distribution is scattered, it benefits from the AIR to ensure the smoothness of the model, while the extended property amplifies it to global consistency within the same class, allowing the limited structural information captured from the ALR to be propagated among subpopulations. 
        According to the comparison of results, we conclude that employing a regularization strategy suitable for data property is important in capturing semantic information. Without local consistency, outlier samples are difficult to learn, which makes the model heavily dependent on the transferability of source knowledge. Similarly, removing input consistency makes it difficult to effectively facilitate the learning of global semantic information. 
        Overall, CtO increased baseline AaD by an average of 2.4\%. This shows that there is complementarity between ALR and AIR.

        \paragraph{Visualization} 
        To demonstrate the superiority of our method, we show the t-SNE~\cite{tsne} feature visualization and confusion matrix on Office-31 (see Fig.~\ref{vis:tsne_con}). 
        From Fig.~\ref{vis:tsne_con}(a-d), we can observe that the clustering of the target features is more compact after the adaptation by CtO. 
        Fig.~\ref{vis:tsne_con}(b) and (d) illustrate that CtO can achieve good model adaptation whether the model is pre-trained on a large-scale or small-scale source domain. 
        When the source domain is knowledge-rich, as shown in Fig.~\ref{vis:a_w_res}, the target domain features already possess considerable semantics. In such cases, adaptive local-consistency regularization can effectively capture the intra-class structure.
        However, when significant domain differences exist (as shown in Fig.~\ref{vis:d_a_res}), abundant target features are jumbled together, so that the model has difficult in capturing the local structure. 
        The flexible data division of our method, thus, customizes the learning strategy for different data properties, which facilitates the estimation of ground-truth structural information instead of only adjusting the neighbor weights as in NRC~\cite{NRC}. 
        Benefiting from the adaptive input-consistency regularization, we can capture semantic structures with rich intra-class variations while dissimilar samples are naturally separated in the representation space. More importantly, as the training iterates, outlier samples gradually join the clustering, and locally informed clusters propagate label information to these outliers. 
        The comparison of Fig.~\ref{vis:tsne_con}(e) and (f) further demonstrates that our method increases prediction diversity by adaptively adjusting the training on under-learned or hard-to-learn samples (\ie outlier).
        
\section{Conclusions}

    In this paper, we propose a novel approach called Chaos to Order (CtO), which tries to achieve efficient feature clustering from the perspective of label propagation. CtO divides the target data into inner and outlier samples based on the adaptive threshold of the learning state, and applies a customized learning strategy to fit the data properties best. To mitigate the source bias, on the one hand, considering the clustering affinity, we propose Adaptive Local-consistency Regularization (ALR) to reduce spurious clustering by re-weighting neighbors. On the other hand, Adaptive Input-consistency Regularization (AIR) is used at outlier points to propagate structural information from high-density to low-density regions, thus achieving high accuracy with respect to the ground truth labels. Moreover, this co-training process can encourage positive clustering and combat spurious clustering. The experimental results of three popular benchmarks verify that our proposed model outperforms the state-of-the-art in various SFDA tasks. 
    For future work, we plan to extend our CtO method to source-free open-set and partial-set domain adaptation.

\begin{acks}
This work was supported by the National Natural Science Foundation of China under Grant 61871186 and 61771322.
\end{acks}

\bibliographystyle{ACM-Reference-Format}
\balance
\bibliography{sample-base}


\begin{thebibliography}{39}


\ifx \showCODEN    \undefined \def \showCODEN     #1{\unskip}     \fi
\ifx \showDOI      \undefined \def \showDOI       #1{#1}\fi
\ifx \showISBNx    \undefined \def \showISBNx     #1{\unskip}     \fi
\ifx \showISBNxiii \undefined \def \showISBNxiii  #1{\unskip}     \fi
\ifx \showISSN     \undefined \def \showISSN      #1{\unskip}     \fi
\ifx \showLCCN     \undefined \def \showLCCN      #1{\unskip}     \fi
\ifx \shownote     \undefined \def \shownote      #1{#1}          \fi
\ifx \showarticletitle \undefined \def \showarticletitle #1{#1}   \fi
\ifx \showURL      \undefined \def \showURL       {\relax}        \fi
\providecommand\bibfield[2]{#2}
\providecommand\bibinfo[2]{#2}
\providecommand\natexlab[1]{#1}
\providecommand\showeprint[2][]{arXiv:#2}

\bibitem[Cai et~al\mbox{.}(2021)]%
        {CaiGLL21}
\bibfield{author}{\bibinfo{person}{Tianle Cai}, \bibinfo{person}{Ruiqi Gao},
  \bibinfo{person}{Jason~D. Lee}, {and} \bibinfo{person}{Qi Lei}.}
  \bibinfo{year}{2021}\natexlab{}.
\newblock \showarticletitle{A Theory of Label Propagation for Subpopulation
  Shift}. In \bibinfo{booktitle}{\emph{{ICML}}}
  \emph{(\bibinfo{series}{Proceedings of Machine Learning Research},
  Vol.~\bibinfo{volume}{139})}. \bibinfo{publisher}{{PMLR}},
  \bibinfo{pages}{1170--1182}.
\newblock


\bibitem[Cubuk et~al\mbox{.}(2020)]%
        {RandAugment}
\bibfield{author}{\bibinfo{person}{Ekin~Dogus Cubuk}, \bibinfo{person}{Barret
  Zoph}, \bibinfo{person}{Jonathon Shlens}, {and} \bibinfo{person}{Quoc Le}.}
  \bibinfo{year}{2020}\natexlab{}.
\newblock \showarticletitle{RandAugment: Practical Automated Data Augmentation
  with a Reduced Search Space}. In \bibinfo{booktitle}{\emph{NeurIPS}}.
\newblock


\bibitem[Ding et~al\mbox{.}(2022)]%
        {SFDADE}
\bibfield{author}{\bibinfo{person}{Ning Ding}, \bibinfo{person}{Yixing Xu},
  \bibinfo{person}{Yehui Tang}, \bibinfo{person}{Chao Xu},
  \bibinfo{person}{Yunhe Wang}, {and} \bibinfo{person}{Dacheng Tao}.}
  \bibinfo{year}{2022}\natexlab{}.
\newblock \showarticletitle{Source-Free Domain Adaptation via Distribution
  Estimation}. In \bibinfo{booktitle}{\emph{{CVPR}}}.
  \bibinfo{publisher}{{IEEE}}, \bibinfo{pages}{7202--7212}.
\newblock


\bibitem[Douze et~al\mbox{.}(2018)]%
        {DouzeSHJ18}
\bibfield{author}{\bibinfo{person}{Matthijs Douze}, \bibinfo{person}{Arthur
  Szlam}, \bibinfo{person}{Bharath Hariharan}, {and}
  \bibinfo{person}{Herv{\'{e}} J{\'{e}}gou}.} \bibinfo{year}{2018}\natexlab{}.
\newblock \showarticletitle{Low-Shot Learning With Large-Scale Diffusion}. In
  \bibinfo{booktitle}{\emph{{CVPR}}}. \bibinfo{publisher}{Computer Vision
  Foundation / {IEEE} Computer Society}, \bibinfo{pages}{3349--3358}.
\newblock


\bibitem[Ganin et~al\mbox{.}(2016)]%
        {DANN}
\bibfield{author}{\bibinfo{person}{Yaroslav Ganin}, \bibinfo{person}{Evgeniya
  Ustinova}, \bibinfo{person}{Hana Ajakan}, \bibinfo{person}{Pascal Germain},
  \bibinfo{person}{Hugo Larochelle}, \bibinfo{person}{Fran{\c{c}}ois
  Laviolette}, \bibinfo{person}{Mario Marchand}, {and}
  \bibinfo{person}{Victor~S. Lempitsky}.} \bibinfo{year}{2016}\natexlab{}.
\newblock \showarticletitle{Domain-Adversarial Training of Neural Networks}.
\newblock \bibinfo{journal}{\emph{J. Mach. Learn. Res.}}  \bibinfo{volume}{17}
  (\bibinfo{year}{2016}), \bibinfo{pages}{59:1--59:35}.
\newblock


\bibitem[He et~al\mbox{.}(2016)]%
        {ResNet}
\bibfield{author}{\bibinfo{person}{Kaiming He}, \bibinfo{person}{Xiangyu
  Zhang}, \bibinfo{person}{Shaoqing Ren}, {and} \bibinfo{person}{Jian Sun}.}
  \bibinfo{year}{2016}\natexlab{}.
\newblock \showarticletitle{Deep Residual Learning for Image Recognition}. In
  \bibinfo{booktitle}{\emph{{CVPR}}}. \bibinfo{publisher}{{IEEE} Computer
  Society}, \bibinfo{pages}{770--778}.
\newblock


\bibitem[Huang et~al\mbox{.}(2021)]%
        {HCL}
\bibfield{author}{\bibinfo{person}{Jiaxing Huang}, \bibinfo{person}{Dayan
  Guan}, \bibinfo{person}{Aoran Xiao}, {and} \bibinfo{person}{Shijian Lu}.}
  \bibinfo{year}{2021}\natexlab{}.
\newblock \showarticletitle{Model Adaptation: Historical Contrastive Learning
  for Unsupervised Domain Adaptation without Source Data}. In
  \bibinfo{booktitle}{\emph{NeurIPS}}. \bibinfo{pages}{3635--3649}.
\newblock


\bibitem[Iscen et~al\mbox{.}(2019)]%
        {IscenTAC19}
\bibfield{author}{\bibinfo{person}{Ahmet Iscen}, \bibinfo{person}{Giorgos
  Tolias}, \bibinfo{person}{Yannis Avrithis}, {and} \bibinfo{person}{Ondrej
  Chum}.} \bibinfo{year}{2019}\natexlab{}.
\newblock \showarticletitle{Label Propagation for Deep Semi-Supervised
  Learning}. In \bibinfo{booktitle}{\emph{{CVPR}}}.
  \bibinfo{publisher}{Computer Vision Foundation / {IEEE}},
  \bibinfo{pages}{5070--5079}.
\newblock


\bibitem[Jin et~al\mbox{.}(2020)]%
        {MCC}
\bibfield{author}{\bibinfo{person}{Ying Jin}, \bibinfo{person}{Ximei Wang},
  \bibinfo{person}{Mingsheng Long}, {and} \bibinfo{person}{Jianmin Wang}.}
  \bibinfo{year}{2020}\natexlab{}.
\newblock \showarticletitle{Minimum Class Confusion for Versatile Domain
  Adaptation}. In \bibinfo{booktitle}{\emph{{ECCV} {(21)}}}
  \emph{(\bibinfo{series}{Lecture Notes in Computer Science},
  Vol.~\bibinfo{volume}{12366})}. \bibinfo{publisher}{Springer},
  \bibinfo{pages}{464--480}.
\newblock


\bibitem[Kang et~al\mbox{.}(2019)]%
        {CAN}
\bibfield{author}{\bibinfo{person}{Guoliang Kang}, \bibinfo{person}{Lu Jiang},
  \bibinfo{person}{Yi Yang}, {and} \bibinfo{person}{Alexander~G. Hauptmann}.}
  \bibinfo{year}{2019}\natexlab{}.
\newblock \showarticletitle{Contrastive Adaptation Network for Unsupervised
  Domain Adaptation}. In \bibinfo{booktitle}{\emph{{CVPR}}}.
  \bibinfo{publisher}{Computer Vision Foundation / {IEEE}},
  \bibinfo{pages}{4893--4902}.
\newblock


\bibitem[Kundu et~al\mbox{.}(2022)]%
        {featmixup}
\bibfield{author}{\bibinfo{person}{Jogendra~Nath Kundu},
  \bibinfo{person}{Akshay~R. Kulkarni}, \bibinfo{person}{Suvaansh Bhambri},
  \bibinfo{person}{Deepesh Mehta}, \bibinfo{person}{Shreyas~Anand Kulkarni},
  \bibinfo{person}{Varun Jampani}, {and} \bibinfo{person}{Venkatesh~Babu
  Radhakrishnan}.} \bibinfo{year}{2022}\natexlab{}.
\newblock \showarticletitle{Balancing Discriminability and Transferability for
  Source-Free Domain Adaptation}. In \bibinfo{booktitle}{\emph{{ICML}}}
  \emph{(\bibinfo{series}{Proceedings of Machine Learning Research},
  Vol.~\bibinfo{volume}{162})}. \bibinfo{publisher}{{PMLR}},
  \bibinfo{pages}{11710--11728}.
\newblock


\bibitem[Kundu et~al\mbox{.}(2021)]%
        {cPAE}
\bibfield{author}{\bibinfo{person}{Jogendra~Nath Kundu},
  \bibinfo{person}{Akshay~R. Kulkarni}, \bibinfo{person}{Amit Singh},
  \bibinfo{person}{Varun Jampani}, {and} \bibinfo{person}{R.~Venkatesh Babu}.}
  \bibinfo{year}{2021}\natexlab{}.
\newblock \showarticletitle{Generalize then Adapt: Source-Free Domain Adaptive
  Semantic Segmentation}. In \bibinfo{booktitle}{\emph{{ICCV}}}.
  \bibinfo{publisher}{{IEEE}}, \bibinfo{pages}{7026--7036}.
\newblock


\bibitem[Lee et~al\mbox{.}(2022)]%
        {JMDS}
\bibfield{author}{\bibinfo{person}{Jonghyun Lee}, \bibinfo{person}{Dahuin
  Jung}, \bibinfo{person}{Junho Yim}, {and} \bibinfo{person}{Sungroh Yoon}.}
  \bibinfo{year}{2022}\natexlab{}.
\newblock \showarticletitle{Confidence Score for Source-Free Unsupervised
  Domain Adaptation}. In \bibinfo{booktitle}{\emph{{ICML}}}
  \emph{(\bibinfo{series}{Proceedings of Machine Learning Research},
  Vol.~\bibinfo{volume}{162})}. \bibinfo{publisher}{{PMLR}},
  \bibinfo{pages}{12365--12377}.
\newblock


\bibitem[Li et~al\mbox{.}(2022)]%
        {AAA}
\bibfield{author}{\bibinfo{person}{Jingjing Li}, \bibinfo{person}{Zhekai Du},
  \bibinfo{person}{Lei Zhu}, \bibinfo{person}{Zhengming Ding},
  \bibinfo{person}{Ke Lu}, {and} \bibinfo{person}{Heng~Tao Shen}.}
  \bibinfo{year}{2022}\natexlab{}.
\newblock \showarticletitle{Divergence-Agnostic Unsupervised Domain Adaptation
  by Adversarial Attacks}.
\newblock \bibinfo{journal}{\emph{{IEEE} Trans. Pattern Anal. Mach. Intell.}}
  \bibinfo{volume}{44}, \bibinfo{number}{11} (\bibinfo{year}{2022}),
  \bibinfo{pages}{8196--8211}.
\newblock


\bibitem[Li et~al\mbox{.}(2020a)]%
        {LiCWW20}
\bibfield{author}{\bibinfo{person}{Rui Li}, \bibinfo{person}{Wenming Cao},
  \bibinfo{person}{Si Wu}, {and} \bibinfo{person}{Hau{-}San Wong}.}
  \bibinfo{year}{2020}\natexlab{a}.
\newblock \showarticletitle{Generating Target Image-Label Pairs for
  Unsupervised Domain Adaptation}.
\newblock \bibinfo{journal}{\emph{{IEEE} Trans. Image Process.}}
  \bibinfo{volume}{29} (\bibinfo{year}{2020}), \bibinfo{pages}{7997--8011}.
\newblock


\bibitem[Li et~al\mbox{.}(2020b)]%
        {3CGAN}
\bibfield{author}{\bibinfo{person}{Rui Li}, \bibinfo{person}{Qianfen Jiao},
  \bibinfo{person}{Wenming Cao}, \bibinfo{person}{Hau{-}San Wong}, {and}
  \bibinfo{person}{Si Wu}.} \bibinfo{year}{2020}\natexlab{b}.
\newblock \showarticletitle{Model Adaptation: Unsupervised Domain Adaptation
  Without Source Data}. In \bibinfo{booktitle}{\emph{{CVPR}}}.
  \bibinfo{publisher}{Computer Vision Foundation / {IEEE}},
  \bibinfo{pages}{9638--9647}.
\newblock


\bibitem[Liang et~al\mbox{.}(2020)]%
        {SHOT}
\bibfield{author}{\bibinfo{person}{Jian Liang}, \bibinfo{person}{Dapeng Hu},
  {and} \bibinfo{person}{Jiashi Feng}.} \bibinfo{year}{2020}\natexlab{}.
\newblock \showarticletitle{Do We Really Need to Access the Source Data? Source
  Hypothesis Transfer for Unsupervised Domain Adaptation}. In
  \bibinfo{booktitle}{\emph{{ICML}}} \emph{(\bibinfo{series}{Proceedings of
  Machine Learning Research}, Vol.~\bibinfo{volume}{119})}.
  \bibinfo{publisher}{{PMLR}}, \bibinfo{pages}{6028--6039}.
\newblock


\bibitem[Long et~al\mbox{.}(2018)]%
        {CDAN}
\bibfield{author}{\bibinfo{person}{Mingsheng Long}, \bibinfo{person}{Zhangjie
  Cao}, \bibinfo{person}{Jianmin Wang}, {and} \bibinfo{person}{Michael~I.
  Jordan}.} \bibinfo{year}{2018}\natexlab{}.
\newblock \showarticletitle{Conditional Adversarial Domain Adaptation}. In
  \bibinfo{booktitle}{\emph{NeurIPS}}. \bibinfo{pages}{1647--1657}.
\newblock


\bibitem[Na et~al\mbox{.}(2021)]%
        {FixBi}
\bibfield{author}{\bibinfo{person}{Jaemin Na}, \bibinfo{person}{Heechul Jung},
  \bibinfo{person}{Hyung~Jin Chang}, {and} \bibinfo{person}{Wonjun Hwang}.}
  \bibinfo{year}{2021}\natexlab{}.
\newblock \showarticletitle{FixBi: Bridging Domain Spaces for Unsupervised
  Domain Adaptation}. In \bibinfo{booktitle}{\emph{{CVPR}}}.
  \bibinfo{publisher}{Computer Vision Foundation / {IEEE}},
  \bibinfo{pages}{1094--1103}.
\newblock


\bibitem[Pei et~al\mbox{.}(2023)]%
        {UITR}
\bibfield{author}{\bibinfo{person}{Jiangbo Pei}, \bibinfo{person}{Zhuqing
  Jiang}, \bibinfo{person}{Aidong Men}, \bibinfo{person}{Liang Chen},
  \bibinfo{person}{Yang Liu}, {and} \bibinfo{person}{Qingchao Chen}.}
  \bibinfo{year}{2023}\natexlab{}.
\newblock \showarticletitle{Uncertainty-Induced Transferability Representation
  for Source-Free Unsupervised Domain Adaptation}.
\newblock \bibinfo{journal}{\emph{{IEEE} Trans. Image Process.}}
  \bibinfo{volume}{32} (\bibinfo{year}{2023}), \bibinfo{pages}{2033--2048}.
\newblock


\bibitem[Peng et~al\mbox{.}(2017)]%
        {VisDA}
\bibfield{author}{\bibinfo{person}{Xingchao Peng}, \bibinfo{person}{Ben Usman},
  \bibinfo{person}{Neela Kaushik}, \bibinfo{person}{Judy Hoffman},
  \bibinfo{person}{Dequan Wang}, {and} \bibinfo{person}{Kate Saenko}.}
  \bibinfo{year}{2017}\natexlab{}.
\newblock \showarticletitle{VisDA: The Visual Domain Adaptation Challenge}.
\newblock \bibinfo{journal}{\emph{CoRR}}  \bibinfo{volume}{abs/1710.06924}
  (\bibinfo{year}{2017}).
\newblock


\bibitem[Qiu et~al\mbox{.}(2021)]%
        {CPGA}
\bibfield{author}{\bibinfo{person}{Zhen Qiu}, \bibinfo{person}{Yifan Zhang},
  \bibinfo{person}{Hongbin Lin}, \bibinfo{person}{Shuaicheng Niu},
  \bibinfo{person}{Yanxia Liu}, \bibinfo{person}{Qing Du}, {and}
  \bibinfo{person}{Mingkui Tan}.} \bibinfo{year}{2021}\natexlab{}.
\newblock \showarticletitle{Source-free Domain Adaptation via Avatar Prototype
  Generation and Adaptation}. In \bibinfo{booktitle}{\emph{{IJCAI}}}.
  \bibinfo{publisher}{ijcai.org}, \bibinfo{pages}{2921--2927}.
\newblock


\bibitem[Qu et~al\mbox{.}(2022)]%
        {BMD}
\bibfield{author}{\bibinfo{person}{Sanqing Qu}, \bibinfo{person}{Guang Chen},
  \bibinfo{person}{Jing Zhang}, \bibinfo{person}{Zhijun Li},
  \bibinfo{person}{Wei He}, {and} \bibinfo{person}{Dacheng Tao}.}
  \bibinfo{year}{2022}\natexlab{}.
\newblock \showarticletitle{{BMD:} {A} General Class-Balanced Multicentric
  Dynamic Prototype Strategy for Source-Free Domain Adaptation}. In
  \bibinfo{booktitle}{\emph{{ECCV} {(34)}}} \emph{(\bibinfo{series}{Lecture
  Notes in Computer Science}, Vol.~\bibinfo{volume}{13694})}.
  \bibinfo{publisher}{Springer}, \bibinfo{pages}{165--182}.
\newblock


\bibitem[Saenko et~al\mbox{.}(2010)]%
        {Office_31}
\bibfield{author}{\bibinfo{person}{Kate Saenko}, \bibinfo{person}{Brian Kulis},
  {et~al\mbox{.}}} \bibinfo{year}{2010}\natexlab{}.
\newblock \showarticletitle{Adapting Visual Category Models to New Domains}. In
  \bibinfo{booktitle}{\emph{ECCV}}.
\newblock


\bibitem[Tang et~al\mbox{.}(2020)]%
        {SRDC}
\bibfield{author}{\bibinfo{person}{Hui Tang}, \bibinfo{person}{Ke Chen}, {and}
  \bibinfo{person}{Kui Jia}.} \bibinfo{year}{2020}\natexlab{}.
\newblock \showarticletitle{Unsupervised Domain Adaptation via Structurally
  Regularized Deep Clustering}. In \bibinfo{booktitle}{\emph{{CVPR}}}.
  \bibinfo{publisher}{Computer Vision Foundation / {IEEE}},
  \bibinfo{pages}{8722--8732}.
\newblock


\bibitem[Van~der Maaten and Hinton(2008)]%
        {tsne}
\bibfield{author}{\bibinfo{person}{Laurens Van~der Maaten} {and}
  \bibinfo{person}{Geoffrey Hinton}.} \bibinfo{year}{2008}\natexlab{}.
\newblock \showarticletitle{Visualizing data using t-SNE.}
\newblock \bibinfo{journal}{\emph{Journal of machine learning research}}
  \bibinfo{volume}{9}, \bibinfo{number}{11} (\bibinfo{year}{2008}).
\newblock


\bibitem[Venkateswara et~al\mbox{.}(2017)]%
        {Office_Home}
\bibfield{author}{\bibinfo{person}{Hemanth Venkateswara}, \bibinfo{person}{Jose
  Eusebio}, \bibinfo{person}{Shayok Chakraborty}, {and}
  \bibinfo{person}{Sethuraman Panchanathan}.} \bibinfo{year}{2017}\natexlab{}.
\newblock \showarticletitle{Deep Hashing Network for Unsupervised Domain
  Adaptation}. In \bibinfo{booktitle}{\emph{{CVPR}}}.
  \bibinfo{publisher}{{IEEE} Computer Society}, \bibinfo{pages}{5385--5394}.
\newblock


\bibitem[Wei et~al\mbox{.}(2021)]%
        {WeiSCM21}
\bibfield{author}{\bibinfo{person}{Colin Wei}, \bibinfo{person}{Kendrick Shen},
  \bibinfo{person}{Yining Chen}, {and} \bibinfo{person}{Tengyu Ma}.}
  \bibinfo{year}{2021}\natexlab{}.
\newblock \showarticletitle{Theoretical Analysis of Self-Training with Deep
  Networks on Unlabeled Data}. In \bibinfo{booktitle}{\emph{{ICLR}}}.
  \bibinfo{publisher}{OpenReview.net}.
\newblock


\bibitem[Wu et~al\mbox{.}(2020)]%
        {DMRL}
\bibfield{author}{\bibinfo{person}{Yuan Wu}, \bibinfo{person}{Diana Inkpen},
  {and} \bibinfo{person}{Ahmed El{-}Roby}.} \bibinfo{year}{2020}\natexlab{}.
\newblock \showarticletitle{Dual Mixup Regularized Learning for Adversarial
  Domain Adaptation}. In \bibinfo{booktitle}{\emph{{ECCV} {(29)}}}
  \emph{(\bibinfo{series}{Lecture Notes in Computer Science},
  Vol.~\bibinfo{volume}{12374})}. \bibinfo{publisher}{Springer},
  \bibinfo{pages}{540--555}.
\newblock


\bibitem[Xia et~al\mbox{.}(2021)]%
        {A2Net}
\bibfield{author}{\bibinfo{person}{Haifeng Xia}, \bibinfo{person}{Handong
  Zhao}, {and} \bibinfo{person}{Zhengming Ding}.}
  \bibinfo{year}{2021}\natexlab{}.
\newblock \showarticletitle{Adaptive Adversarial Network for Source-free Domain
  Adaptation}. In \bibinfo{booktitle}{\emph{{ICCV}}}.
  \bibinfo{publisher}{{IEEE}}, \bibinfo{pages}{8990--8999}.
\newblock


\bibitem[Xu et~al\mbox{.}(2019)]%
        {SAFN}
\bibfield{author}{\bibinfo{person}{Ruijia Xu}, \bibinfo{person}{Guanbin Li},
  \bibinfo{person}{Jihan Yang}, {and} \bibinfo{person}{Liang Lin}.}
  \bibinfo{year}{2019}\natexlab{}.
\newblock \showarticletitle{Larger Norm More Transferable: An Adaptive Feature
  Norm Approach for Unsupervised Domain Adaptation}. In
  \bibinfo{booktitle}{\emph{{ICCV}}}. \bibinfo{publisher}{{IEEE}},
  \bibinfo{pages}{1426--1435}.
\newblock


\bibitem[Yang et~al\mbox{.}(2021a)]%
        {NRC}
\bibfield{author}{\bibinfo{person}{Shiqi Yang}, \bibinfo{person}{Yaxing Wang},
  \bibinfo{person}{Joost van~de Weijer}, \bibinfo{person}{Luis Herranz}, {and}
  \bibinfo{person}{Shangling Jui}.} \bibinfo{year}{2021}\natexlab{a}.
\newblock \showarticletitle{Exploiting the Intrinsic Neighborhood Structure for
  Source-free Domain Adaptation}. In \bibinfo{booktitle}{\emph{NeurIPS}}.
  \bibinfo{pages}{29393--29405}.
\newblock


\bibitem[Yang et~al\mbox{.}(2021b)]%
        {GSFDA}
\bibfield{author}{\bibinfo{person}{Shiqi Yang}, \bibinfo{person}{Yaxing Wang},
  \bibinfo{person}{Joost van~de Weijer}, \bibinfo{person}{Luis Herranz}, {and}
  \bibinfo{person}{Shangling Jui}.} \bibinfo{year}{2021}\natexlab{b}.
\newblock \showarticletitle{Generalized Source-free Domain Adaptation}. In
  \bibinfo{booktitle}{\emph{{ICCV}}}. \bibinfo{publisher}{{IEEE}},
  \bibinfo{pages}{8958--8967}.
\newblock


\bibitem[Yang et~al\mbox{.}(2022)]%
        {AaD}
\bibfield{author}{\bibinfo{person}{Shiqi Yang}, \bibinfo{person}{Yaxing Wang},
  \bibinfo{person}{Kai Wang}, \bibinfo{person}{Shangling Jui}, {et~al\mbox{.}}}
  \bibinfo{year}{2022}\natexlab{}.
\newblock \showarticletitle{Attracting and dispersing: A simple approach for
  source-free domain adaptation}. In \bibinfo{booktitle}{\emph{Advances in
  Neural Information Processing Systems}}.
\newblock


\bibitem[Yi et~al\mbox{.}(2023)]%
        {NRC_ELR}
\bibfield{author}{\bibinfo{person}{Li Yi}, \bibinfo{person}{Gezheng Xu},
  \bibinfo{person}{Pengcheng Xu}, \bibinfo{person}{Jiaqi Li},
  \bibinfo{person}{Ruizhi Pu}, \bibinfo{person}{Charles Ling},
  \bibinfo{person}{A.~Ian McLeod}, {and} \bibinfo{person}{Boyu Wang}.}
  \bibinfo{year}{2023}\natexlab{}.
\newblock \showarticletitle{When Source-Free Domain Adaptation Meets Learning
  with Noisy Labels}. In \bibinfo{booktitle}{\emph{{ICLR}}}.
  \bibinfo{publisher}{OpenReview.net}.
\newblock


\bibitem[Zhang et~al\mbox{.}(2021)]%
        {FlexMatch}
\bibfield{author}{\bibinfo{person}{Bowen Zhang}, \bibinfo{person}{Yidong Wang},
  \bibinfo{person}{Wenxin Hou}, \bibinfo{person}{Hao Wu},
  \bibinfo{person}{Jindong Wang}, \bibinfo{person}{Manabu Okumura}, {and}
  \bibinfo{person}{Takahiro Shinozaki}.} \bibinfo{year}{2021}\natexlab{}.
\newblock \showarticletitle{FlexMatch: Boosting Semi-Supervised Learning with
  Curriculum Pseudo Labeling}. In \bibinfo{booktitle}{\emph{NeurIPS}}.
  \bibinfo{pages}{18408--18419}.
\newblock


\bibitem[Zhang et~al\mbox{.}(2019)]%
        {MDD}
\bibfield{author}{\bibinfo{person}{Yuchen Zhang}, \bibinfo{person}{Tianle Liu},
  \bibinfo{person}{Mingsheng Long}, {and} \bibinfo{person}{Michael~I. Jordan}.}
  \bibinfo{year}{2019}\natexlab{}.
\newblock \showarticletitle{Bridging Theory and Algorithm for Domain
  Adaptation}. In \bibinfo{booktitle}{\emph{{ICML}}}
  \emph{(\bibinfo{series}{Proceedings of Machine Learning Research},
  Vol.~\bibinfo{volume}{97})}. \bibinfo{publisher}{{PMLR}},
  \bibinfo{pages}{7404--7413}.
\newblock


\bibitem[Zhang et~al\mbox{.}(2022)]%
        {DaC}
\bibfield{author}{\bibinfo{person}{Ziyi Zhang}, \bibinfo{person}{Weikai Chen},
  \bibinfo{person}{Hui Cheng}, \bibinfo{person}{Zhen Li},
  \bibinfo{person}{Siyuan Li}, \bibinfo{person}{Liang Lin}, {and}
  \bibinfo{person}{Guanbin Li}.} \bibinfo{year}{2022}\natexlab{}.
\newblock \showarticletitle{Divide and Contrast: Source-free Domain Adaptation
  via Adaptive Contrastive Learning}. In \bibinfo{booktitle}{\emph{Advances in
  Neural Information Processing Systems}}.
\newblock


\bibitem[Zhong et~al\mbox{.}(2021)]%
        {E-MixNet}
\bibfield{author}{\bibinfo{person}{Li Zhong}, \bibinfo{person}{Zhen Fang},
  \bibinfo{person}{Feng Liu}, \bibinfo{person}{Jie Lu}, \bibinfo{person}{Bo
  Yuan}, {and} \bibinfo{person}{Guangquan Zhang}.}
  \bibinfo{year}{2021}\natexlab{}.
\newblock \showarticletitle{How Does the Combined Risk Affect the Performance
  of Unsupervised Domain Adaptation Approaches?}. In
  \bibinfo{booktitle}{\emph{{AAAI}}}. \bibinfo{publisher}{{AAAI} Press},
  \bibinfo{pages}{11079--11087}.
\newblock


\end{thebibliography}

\clearpage
\appendix

\section{Proof of Claim~\ref{claim:dis_setting} and Theorem~\ref{thm:bound}}
    \label{app:proof}


    \subsection{Claim~\ref{claim:dis_setting} and Proof}
        
        \textsc{Claim} 3.1.
    	\textit{Suppose $G$ satisfies a Lipschitz condition; there exists a global threshold $\rho \in (0, 1)$ and scale of models' learning status $\tau_i$ such that the inner set $I$ is consistency robust, \ie $R_{\mathcal{B}}(G) = 0$. More specifically, }
            $$\begin{aligned}
                & r \leq (2 \max\{\tau_i\} \rho - 1) \frac{2}{L}, \\
                & \forall i \in [\mathcal{C}].
            \end{aligned}$$
    
        \paragraph{Proof} 
        Let $G(x)_{[k]}$ denote the predicted probability of the model on class $k$. 
        Integrating the dynamic thresholds, we say that there exists a constant $\gamma$ such that $G(x)_{[j]} \in [\tau_j \rho - \gamma, 1 - \tau_i \rho]$, where $j \neq \arg \max (G(x))$. 
        Suppose $I$ is defined by model's learning state, denote 
        $$I \triangleq \{x: \max (G(x)) \geq \tau_i \rho \; s.t.\; i=\arg \max (G(x)) \}.$$
        If $\exists x, x^{\prime} \in I, x^{\prime} \in \mathcal{B}(x) \cap I$ s.t. $G(x) \neq G(x^{\prime})$. Specifically, define
        $$\arg \max G(x) = i, \; \arg \max G(x^{\prime}) = j,$$
        where $i \neq j$. Along with the inequality we know that
        $$|G(x)_{[i]} - G(x^{\prime})_{[j]}| \geq 2\tau_i \rho - \gamma - 1.$$
        By the definition of Lipschitz constant, we have:   
        \begin{equation}
            \begin{aligned}
            \label{eq:lh}
                L\left\|x-x^{\prime}\right\| &\geq \left\|G(x)-G\left(x^{\prime}\right)\right\| \\
                &\geq\left|G(x)_{[i]} + G\left(x\right)_{[j]} - G(x^{\prime})_{[j]} -G\left(x^{\prime}\right)_{[i]}\right| \\
                &\geq\left|G(x)_{[i]}-G\left(x^{\prime}\right)_{[j]}\right|+\left|G(x)_{[j]}-G\left(x^{\prime}\right)_{[i]}\right| \\
                &\geq 4(\tau_i + \tau_j)\rho - 2 - 2 \gamma \\
                &\geq 4 (\tau_i + \tau_j) \rho - 2.
            \end{aligned}
        \end{equation}
        As a result,
        $$\begin{aligned}
            & L\left\|x-x^{\prime}\right\| \geq 4 \max (\tau_i) \rho - 2, \\
            & \forall i \in [\mathcal{C}].
        \end{aligned}$$
        Since by definition of $x^{\prime} \in \mathcal{B}$, we have $\|x - x^{\prime}\| \leq r$. 
        Combining Eq.~\ref{eq:lh} and the Lipschitz constant $L > 0$, we know that this forms a contradiction with $L\left\|x-x^{\prime}\right\| \geq L r$. 
        Thus, $\forall x \in I, x^{\prime} \in \mathcal{B}(x) \cap I$, the model predictions are consistent, \ie $R_{\mathcal{B}}(G) = 0$. 

    \subsection{Proof Sketch for Theorem~\ref{thm:bound}}
            \textsc{Theorem} 3.1.
        	\textit{Suppose Assumption~\ref{asm_setting} and Claim~\ref{claim:dis_setting} hold and $I, O$ satisfies $(q, \mu)$-constant expansion. Then the expected error of model $G$ is bounded,}
                $$\epsilon_{\mathcal{D}_T}(G) \leq 4 \max(q, \mu) \kappa + \mu (1 + \kappa).$$

        To prove the Theorem~\ref{thm:bound}, we introduces some concepts and notations following~\cite{CaiGLL21}: (i) the robust set of $G$, $RS(G)$; (ii) the minority robust set of $G$ on $U$, $M$.
    
        For a given model $G$, define the robust set to be the set for which $G$ is robust under input transformations:
        $$RS(G) := \{x| G(x) = G(x^{\prime}), \forall x^{\prime} \in \mathcal{B}(x)\}.$$

        Let $A_{ik} \triangleq RS(G) \cap U_i \cap\{x|G(x)=k\} \;s.t.\; i, k \in [\mathcal{C}]$, where $U_i$ denote the conditional distribution of $U$. 
        Towards define the minority robust set $M$ on $U$, we consider the majority class label of $G$:
        $$y_i^{\text{Maj}} \triangleq \arg \max_{k\in [\mathcal{C}]}\mathbb{P}_{U}[A_{ik}].$$ 
        Thus, we denote 
        $$M \triangleq \bigcup_{k\in [\mathcal{C}] \backslash \{y_i^{\text{Maj}}\}}A_{ik}$$
        be the minority robust set of $G$.
        In addition, let
        $$\widetilde{M} \triangleq \bigcup_{i\in [\mathcal{C}]} \left(U_i \cap \{x|G(x) \neq y_i^{\text{Maj}}\}\right)$$
        be the minority set of $G$.
    
        By the Lemma A.1 in~\cite{CaiGLL21}, under the $(q, \mu)$-constant expansion, we have $$\begin{aligned}
            &\mathbb{P}_{U}[M] \le 2\max(q, \mu), \\
            &\mathbb{P}_{U}[\widetilde{M}] \le 2\max(q, \mu) + \mu.
        \end{aligned}$$
    
        \begin{lemma}[Upper Bound on the inner set $I$]\label{lem_I}
    	Suppose the condition of Claim~\ref{claim:dis_setting} holds, then
    	$$\epsilon_I(G) \le \mathbb{P}_{I}[M] + R_{\mathcal{B}}(G).$$
        \end{lemma}
    
        \paragraph{Proof} 
        Based on the definition of the minority robust set $M$, we know that $I \triangleq M \cup \{x: G(x) \neq G(x^{\prime}), x \in I, \;\text{and}\; x^{\prime} \in \mathcal{B}(x) \cap O \}$. Therefore, we can write: 
        \begin{equation}
            \begin{aligned}
                \epsilon_I(G) &= \mathbb{P}_I[G(x) \neq G^{*}(x)] \\
                &= \mathbb{P}_I[M \cap (G(x) \neq G^{*}(x))] \\
                &\;\;\;\; + \mathbb{P}_I[(G(x) \neq G(x^{\prime})) \cap (G(x) \neq G^{*}(x))] \\
                &\le \mathbb{P}_I[M] + \mathbb{P}_I[\overline{RS(G)}] \\
                &\le \mathbb{P}_I[M] + R_{\mathcal{B}}(G).
            \end{aligned}
        \end{equation}
        
        \begin{lemma}[Upper Bound on the outlier set $O$]\label{lem_O}
    	Let $O=\mathcal{D}_T \backslash I$, then
    	$$\epsilon_O(G) \le \mathbb{P}_{O}[M] + \mathbb{P}_{O}[\widetilde{M}] + R_{\mathcal{B}}(G).$$
        \end{lemma}
        
        \paragraph{Proof} By the definition of the outlier set $O$, we note that $\{x: G(x) \neq G^*(x), \;\text{and}\; x \in O\} \subseteq M \cup \widetilde{M} \cup (\overline{RS(G)} \backslash \widetilde{M})$. Thus, we obtain
        \begin{equation}
            \begin{aligned}
                \epsilon_O(G) \le \mathbb{P}_{O}[M] + \mathbb{P}_{O}[\widetilde{M}] + R_{\mathcal{B}}.
            \end{aligned}
        \end{equation}
    
        Based on the above results, we can now apply Lemma~\ref{lem_I} and Lemma~\ref{lem_O} to bound the target error $\epsilon_{\mathcal{D}_T}(G)$. 
        Under the conditions of Theorem~\ref{thm:bound}, we have:
    
    
        \begin{equation}
            \begin{aligned}
                \epsilon_{\mathcal{D}_T}(G) &= \mathbb{P}_{\mathcal{D}_T}[I]\epsilon_I(G) + \mathbb{P}_{\mathcal{D}_T}[O]\epsilon_O(G) \\
                &\le \mathbb{P}_{\mathcal{D}_T}[I]\left( \mathbb{P}_{I}[M] + R_{\mathcal{B}}(G) \right) \\
                &\;\;\;\; + \mathbb{P}_{\mathcal{D}_T}[O](\mathbb{P}_{O}[M] + \mathbb{P}_{O}[\widetilde{M}] + R_{\mathcal{B}}(G)) \\
                & \text{(Lemma~\ref{lem_I} and Lemma~\ref{lem_O})} \\
                &\le \mathbb{P}_{\mathcal{D}_T}[I]\left( \kappa \mathbb{P}_{U}[M] + R_{\mathcal{B}}(G) \right) \\
                &\;\;\;\; + \mathbb{P}_{\mathcal{D}_T}[O]( \kappa \mathbb{P}_{U}[M] + \kappa \mathbb{P}_{U}[\widetilde{M}] + R_{\mathcal{B}}(G)) \\
                & \text{(Assumption~\ref{asm_setting})} \\
                &\le \kappa \mathbb{P}_{U}[M] + R_{\mathcal{B}}(G) + \kappa \mathbb{P}_{\mathcal{D}_T}[O] \mathbb{P}_{U}[\widetilde{M}] \\
                &\le 4 \max(q, \mu) \kappa + \mu (1 + \kappa).
            \end{aligned}
        \end{equation}

\begin{table*}[t]
    \begin{center}
        \setlength\tabcolsep{3pt}
        \scalebox{0.9}{  
            \begin{tabular}{c|cccccccccccccc}
                \toprule
                \multicolumn{2}{c}{Layer} & Ar$\rightarrow$Cl & Ar$\rightarrow$Pr & Ar$\rightarrow$Rw & Cl$\rightarrow$Ar & Cl$\rightarrow$Pr & Cl$\rightarrow$Rw & Pr$\rightarrow$Ar & Pr$\rightarrow$Cl & Pr$\rightarrow$Rw & Rw$\rightarrow$Ar & Rw$\rightarrow$Cl & Rw$\rightarrow$Pr & Avg. \\
                \midrule
                \multirow{2}{1.7cm}{Layer 4 (source)} 					   & same class & 14.848 & 12.155 & 12.918 & 13.849 & 11.914 & 12.650 & 16.183 & 17.240 & 14.453 & 14.183 & 13.989 & 11.680 & 13.839 \\
				& across classes & 16.674 & 15.556 & 15.548 & 15.449 & 14.317 & 14.597 & 18.152 & 18.730 & 16.924 & 15.978 & 15.410 & 14.548 & 15.990 \\
				\midrule
				\multirow{2}{1.7cm}{Layer 4 (target)} 					   & same class & 12.800 & 12.961 & 12.275 & 14.183 & 12.961 & 12.275 & 14.183 & 12.800 & 12.275 & 14.183 & 12.800 & 12.961 & 13.055 \\
				& across classes & 14.422 & 16.174 & 14.750 & 16.396 & 16.174 & 14.750 & 16.396 & 14.422 & 14.750 & 16.396 & 14.422 & 16.174 & 15.435 \\
				\midrule
				\multirow{2}{1.7cm}{Bottleneck (source)}   & same class & 19.041 & 15.302 & 16.117 & 18.614 & 16.453 & 17.260 & 18.656 & 20.286 & 16.868 & 16.799 & 17.428 & 14.102 & 17.244 \\
				& across classes & 21.975 & 20.965 & 21.160 & 21.529 & 20.717 & 20.926 & 21.778 & 22.301 & 21.182 & 21.033 & 20.275 & 20.274 & 21.176 \\
				\midrule
				\multirow{2}{1.7cm}{Bottleneck (target)}    & same class & 19.813 & 13.601 & 14.845 & 16.137 & 13.223 & 14.362 & 18.703 & 22.772 & 16.631 & 16.223 & 18.673 & 12.934 & 16.493 \\
				& across classes & 24.252 & 20.307 & 22.235 & 20.172 & 18.489 & 19.838 & 23.969 & 26.605 & 23.761 & 21.410 & 22.489 & 19.170 & 21.891 \\
                \bottomrule
            \end{tabular}
        }
    \end{center}
    \caption{\label{tab:Euclidean}Euclidean distance within the same class and across classes in each task on \textbf{Office-Home}.}
\end{table*}
    
\section{Algorithm for CtO}
\label{app:CtO}
    Our method, as described in Algorithm~\ref{alg:CtO}, involves dynamic data grouping and adaptive input- and local-consistency regularization. By using an adaptive threshold for the learning state, we are able to dynamically divide the target data into Inner data and outlier data. This allows us to apply different learning strategies to each subset.
    
    \begin{algorithm}[t]
        \caption{Training Algorithm of CtO}
        \label{alg:CtO}
        \begin{algorithmic}[1]
            \renewcommand{\algorithmicrequire}{\textbf{Input:}}
            \REQUIRE{Unlabeled target data $\mathcal{X}_t$, number of classes $C$, a pre-trained source model $G = g_c \circ h$, global threshold $\rho$.}
            \renewcommand{\algorithmicrequire}{\textbf{Initialisation:}}
            \REQUIRE{Initialisation: Build feature bank $\boldsymbol{F}$ and score bank $\boldsymbol{P}$ by forward computation, initialize the global threshold as $1/C$.}
            \WHILE{$t < \text{MaxIterations}$}
                \STATE Update $\boldsymbol{F}$ and $\boldsymbol{P}$ corresponding to the current mini-batch $B$;
                \STATE Update the global threshold based on Eq.~\ref{reweight:gc};
                \STATE Compute the learning effect of class based on Eq.~\ref{eq:le};
                \STATE Update the inner and outlier sample based on Eq.~\ref{eq:division};
                \STATE Compute $\mathcal{L}_{alr}$ on inner sample by Eq.~\ref{loss:alr};
                \STATE Compute $\mathcal{L}_{sep}$ on all sample by Eq.~\ref{loss:sep};
                \STATE Compute $\mathcal{L}_{air}$ on outlier sample by Eq.~\ref{loss:air};
                \STATE Compute overall loss and update the model $G$.
            \ENDWHILE
            \renewcommand{\algorithmicensure}{\textbf{Output:}}
            \ENSURE{The adapted model $G$.}
        \end{algorithmic}
    \end{algorithm}

\section{Analysis of different similarity measures}
    \label{app:different_sim}

\begin{figure}[t]
        \centering
        \includegraphics[width=7.5cm]{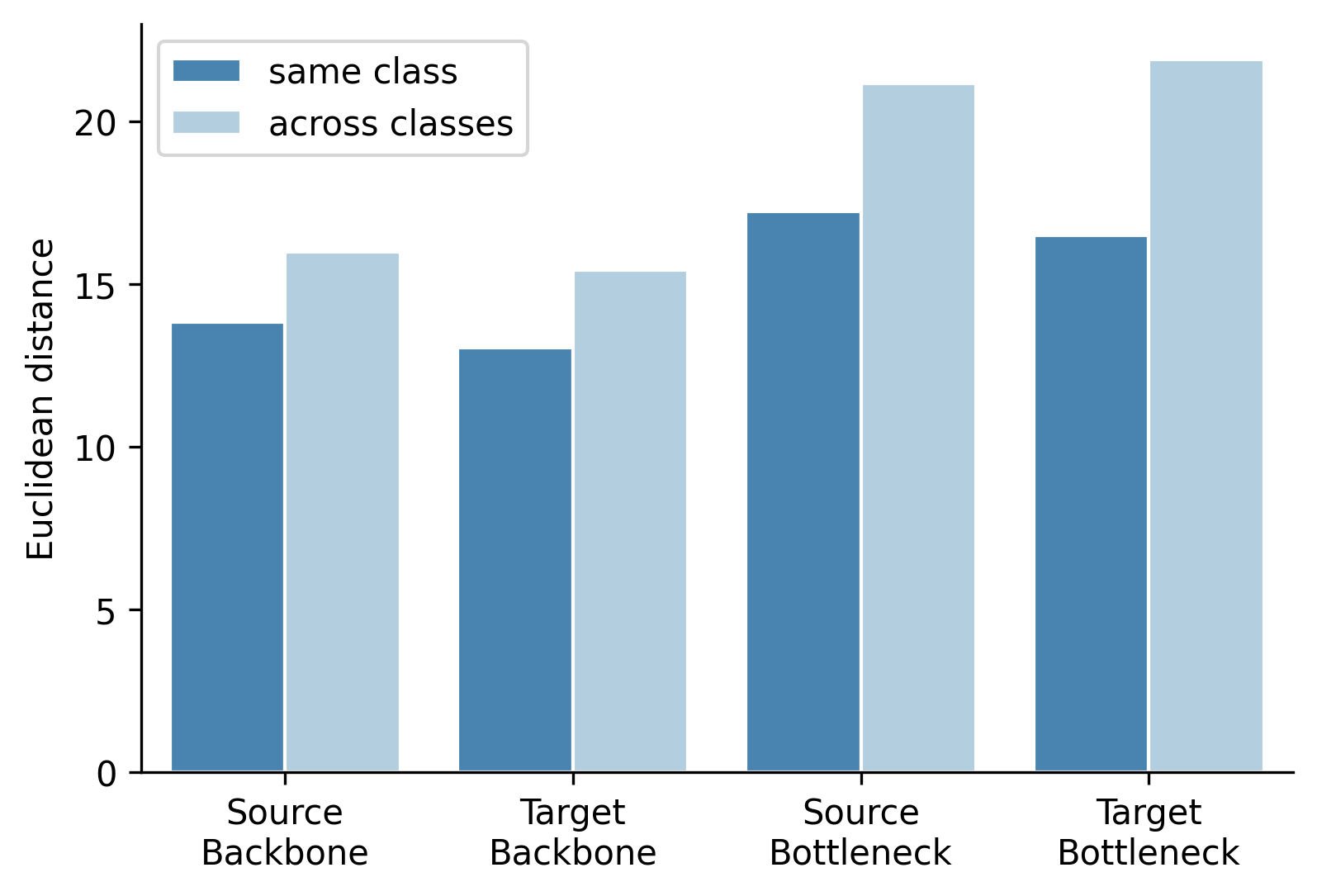}
        \caption{Histogram of the Euclidean distance within the same class and across classes on Office-Home.}
        \label{vis:Euclidean}
    \end{figure}

    To check the effect of different metric functions on target feature clustering, we compared the Cosine similarity and the Euclidean distance on Office-Home for feature similarity. Fig.~\ref{vis:Euclidean} shows the average Euclidean distance for all tasks on Office-Home. A smaller value indicates more similarity. 
    It can be seen that the Euclidean distance also indicates the presence of rich semantic information in the high-dimensional features of the source model backbone network. However, compared to Fig.~\ref{vis:sim}, the differences in similarity based on Euclidean distance are insignificant. 
    As is well known, the Euclidean distance reflects absolute differences in values, while the cosine distance reflects relative differences in direction. 
    Therefore, Cosine similarity maintains "1 for identical, 0 for orthogonal, -1 for opposite" in high-dimensional space. Euclidean distance, in contrast, is influenced by dimensions, and its numerical space is not unstable. 
    Particularly in the case of distribution shifts, the variance of the sample fluctuations is too large, leading to poor Euclidean distance performance.
    
    We also shows feature similarities among samples within the same class and across classes in each transfer task. 
    As shown in Table~\ref{tab:Euclidean}, the differences between the Euclidean distance within the same class and that across classes are not clear if the distributions are significantly different (e.g., Ar$\rightarrow$Cl, Pr$\rightarrow$Cl, Rw$\rightarrow$Cl tasks). Weak inter-category discrimination exacerbates spurious clustering, which biases the adaptation process. 

    Through the experiment, we notice two things: 1) the source model contains sufficient inductive biases; and 2) under domain shift conditions, there is a strong correlation between the metric method and target feature clustering. 
    In future work, one possible direction is to study how different metrics affect the performance of target clustering.

\end{document}